\documentclass[acmsmall]{acmart}

\usepackage{csquotes}
\newcommand{\indep}{\rotatebox[origin=c]{90}{$\models$}}

\graphicspath{ {./images/} }

\def\CircLeftarrow{\kern3pt\hbox{$\leftarrow$}\kern-2pt\hbox{$\circ$}\kern3pt}

\def\CircRightarrow{\kern3pt\hbox{$\circ$}\kern-2pt\hbox{$\rightarrow$}\kern3pt}

\def\CircCirc{\kern3pt\hbox{$\circ$}\kern-1pt\hbox{-}\kern-1pt\hbox{-}\kern-1pt\hbox{$\circ$}\kern3pt}

\def\CircLine{\kern3pt\hbox{$\circ$}\kern-1pt\hbox{-}\kern-1pt\kern-1pt\hbox{-}\kern-1pt\hbox{-}\kern-1pt\hbox{-}\kern3pt}

\def\LineCirc{\kern3pt\hbox{-}\kern-1pt\hbox{-}\kern-1pt\hbox{-}\kern-1pt\hbox{$\circ$}\kern3pt}

\AtBeginDocument{%
  \providecommand\BibTeX{{%
    \normalfont B\kern-0.5em{\scshape i\kern-0.25em b}\kern-0.8em\TeX}}}

\setcopyright{acmcopyright}
\copyrightyear{2021}
\acmYear{2021}
\acmDOI{10.1145/1122445.1122456}

\acmJournal{CSUR}
\acmVolume{}
\acmNumber{}
\acmArticle{}
\acmMonth{}

\begin{document}

\title{D'ya like DAGs? A Survey on Structure Learning and Causal Discovery}

\author{Matthew J. Vowels}
\authornote{Corresponding author.}
\email{m.j.vowels@surrey.ac.uk}
\author{Necati Cihan Camgoz}
\email{n.camgoz@surrey.ac.uk}
\author{Richard Bowden}
\email{r.bowden@surrey.ac.uk}
\affiliation{%
  \institution{CVSSP, University of Surrey}
  \city{Guildford}
  \state{Surrey}
  \postcode{GU2 7XH}
  \country{U.K.}
}

\renewcommand{\shortauthors}{Vowels et al.}

\begin{abstract}
Causal reasoning is a crucial part of science and human intelligence. In order to discover causal relationships from data, we need structure discovery methods. We provide a review of background theory and a survey of methods for structure discovery. We primarily focus on modern methods which leverage continuous optimization, and provide reference to further resources such as benchmark datasets and software packages. Finally, we discuss the assumptive leap required to take us from structure to causality.
\end{abstract}


\keywords{}


\maketitle

\section{Introduction}
Causal understanding has been described as `part of the bedrock of intelligence' \cite{Li2020b}, and is one of the fundamental goals of science \cite{Glymour2019,Pearl2009, vanderLaan2011, Bareinboim2020, vanderLaan2014, vanderLaan2018}. It is important for a broad range of applications, including policy making \cite{Kreif2019}, medical imaging \cite{Castro2019}, advertisement \cite{Bottou2013}, the development of medical treatments \cite{Petersen2017}, the evaluation of evidence within legal frameworks \cite{Pearl2009, Siegerink2017}, social science \cite{Vowels2021, Hernan2018, Grosz2020}, biology \cite{Triantafillou2017}, and many others. It is also a burgeoning topic in machine learning and artificial intelligence \cite{Xu2020, Scholkopf2019, Li2020, Bengio2019b, Goyal2020, Vowels2020b, Garcez2020}, where it has been argued that a consideration for causality is crucial for reasoning about the world. In order to discover causal relations, and thereby gain causal understanding, one may perform interventions and manipulations as part of a randomized experiment. These experiments may not only allow researchers or agents to identify causal relationships, but also to estimate the magnitude of these relationships.

Unfortunately, in many cases, it may not be possible to undertake such experiments due to prohibitive cost, ethical concerns, or impracticality. For example, to understand the impact of smoking, it would be necessary to force different individuals to smoke or not-smoke. Researchers are therefore often left with non-experimental, observational data. In the absence of intervention and manipulation, observational data leave researchers facing a number of challenges: Firstly, observational datasets may not contain all relevant variables - there may exist unobserved/hidden/latent factors (this is sometimes referred to as the third variable problem). Secondly, observational data may exhibit selection bias - for example, younger patients may in general prefer to opt for surgery, whereas older patients may prefer medication. Thirdly, the causal relationships underlying these data may not be known \textit{a priori} - for example, are genetic factors independent causes of a particular outcome, or do they mediate or moderate an outcome? These three challenges affect the discovery and estimation of causal relationships.

To address these challenges, researchers in the fields of statistics and machine learning have developed numerous methods for uncovering causal relations (causal discovery) and estimating the magnitude of these effects (causal inference) from observational data, or from a mixture of observational and experimental data. Under various (often strong) assumptions, these methods are able to take advantage of the relative abundance of observational data in order to infer causal structure and causal effects. Indeed, observational data may, in spite of the three challenges listed above, provide improved statistical power and generalizability compared with experimental data \cite{Deaton2018}.

In this paper we review relevant background theory and provide a survey of methods which perform structure discovery (sometimes called causal induction \cite{Griffiths2009}) with observational data or with a mixture of observational and experimental data. A number of reviews, surveys and guides are already available (see \textit{e.g.} \cite{Heinze2018, Glymour2019, Spirtes2016}), however, these reviews cover combinatoric approaches to causal discovery, whereas we primarily focus on the recent flurry of developments in continuous optimization approaches. Furthermore, the existing reviews are relatively short, and we attempt to provide a more scoping introduction to the necessary background material. We also seek to provide more extensive coverage of continuous optimization approaches than other current reviews, which focus on combinatoric approaches. Finally, we provide references to further useful resources including datasets and openly available software packages.

The structure of this survey is as follows: Following an overview of relevant background information in Section \ref{sec:background}, we provide an overview of approaches to structure discovery in Section \ref{sec:discovery}, including a list of common evaluation metrics. In Section \ref{sec:comb} we briefly outline a range of combinatoric approaches, before focusing on continuous optimization approaches in Section \ref{sec:cont}. We begin \ref{sec:summary} by referencing several additional resources including reviews, guides, datasets, and software packages. We also provide a summary and discussion of the methods covered in Section \ref{sec:summary}, and note various opportunities for future work and future direction. Many of the methods we review in this survey seek to discover and interpret the learned structure \textit{causally}. Whilst this is a laudable aim, we are reminded of important commentaries (\textit{e.g.},\cite{Humphreys1996,Freedman1998,Dawid2008}) which argue for appropriate skepticism and care when making the leap from observation to causality via causal discovery methods. We therefore conclude Section \ref{sec:summary}, as well as this survey as a whole, by providing a discussion on these issues.

\section{Background - Definitions and Assumptions}
\label{sec:background}
In this section we provide working definitions of key concepts in structure discovery. We include a presentation of a common framework used in structure discovery (namely, that of structured graphical representations) as well as a number of common assumptions.

\subsection{Causality and SCMs}
In spite of some notable reluctance to treat graphs learned from observational data as causal \cite{Dawid2008, Humphreys1996, Freedman1998}, we acknowledge that it is a common and worthwhile aim, and begin by presenting a working definition of causality and its popular systematization in Structural Causal Models (SCMs). Causality eludes straightforward definition \cite{Spirtes2000}, and is often characterized intuitively with examples involving fires and houses \cite{Pearl2009}, firing squads \cite{Hopkins2003}, and bottles and rocks \cite{Hopkins2005}. One definition of what is known as \textit{counterfactual causality} is given by by Lewis (1973) \cite{LewisCausation1973} as follows:\footnote{See discussion in Menzies \& Beebee (2020) \cite{MenziesCausation2020}} 

\begin{displayquote}"We think of a cause as something that makes a difference, and the difference it makes must be a difference from what would have happened without it. Had it been absent, its effects – some of them, at least, and usually all – would have been absent as well".
\end{displayquote}

Lewis' definition is counterfactual in the sense that he effectively describes `what would have happened if the cause had been A*, given that the effect was B when the cause was A'. Seemingly, this definition is compatible with the \textit{Pearlian} school of causal reasoning. Specifically, in the context of what are known as SCMs:

\begin{displayquote}
"Given two disjoint sets of variables $X$ and $Y$, the causal effect of $X$ on $Y$, denoted as... $P(y|do(x))$, is a function from $X$ to the space of probability distributions on $Y$. For each realization of $x$ of $X$, $P(y|do(x))$ gives the probability of $Y=y$ induced by deleting from the model [$x_i = f_i(pa_i,u_i), i=1...,n,$] all equations corresponding to variables in $X$ and substituting $X=x$ in the remaining equations."
\end{displayquote}

\begin{figure}[t!]
\centering
\includegraphics[width=0.5\linewidth]{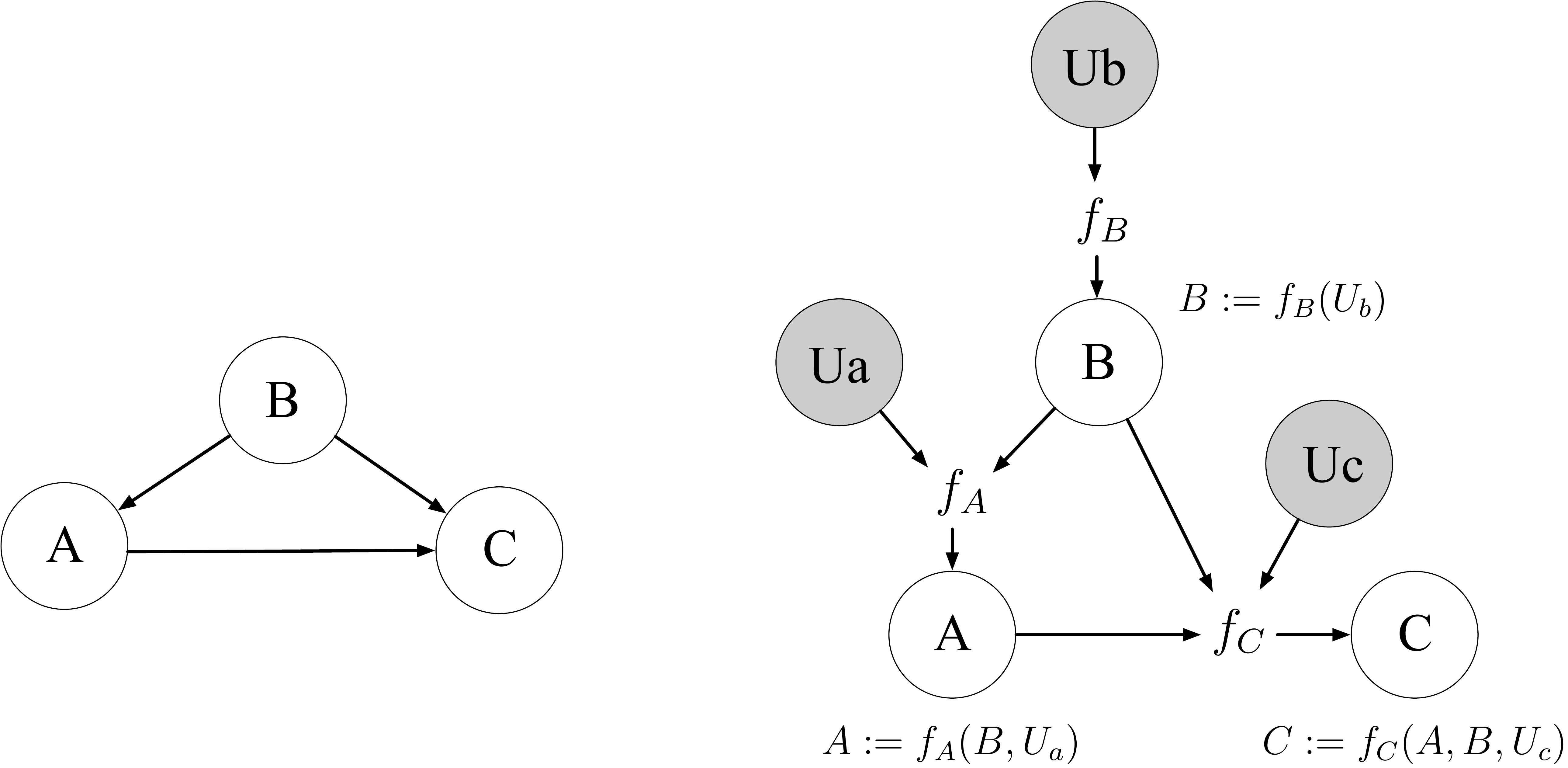}
\caption{Transitioning from a typical DAG representation (left) to a structural equation model (right). Grey vertices are unobserved/latent random variables.}
\label{fig:sem}
\end{figure}

This definition \cite[p.70]{Pearl2009} requires further examination. Firstly, the model $x_i = f_i(pa_i,u_i), i=1...,n,$ is a Structural Equation/Causal Model (SEM/SCM) which indicates assignment of the value $x_i$ in the space of $X$ to a function of its structural parents $pa_i$ and exogenous noise $u_i$. We elaborate on what parents are (as well as children, descendants etc.) below. Secondly, the $do$ notation \cite{Pearl2009} indicates \textit{intervention}, where the value of $x$ is set to a specific quantity. The structure (including attributes such as \textit{parents}) can be represented graphically using various types of graphical models (\textit{e.g.}, Directed Acyclic Graphs). Figure \ref{fig:sem} shows the relationship between a DAG and a general Structural Equation Model. Sometimes this SEM is also called a Functional Causal Model (FCM), where the functions are assumed to represent the causal mechanisms \cite{Goudet2018}. The use of the assignment operator `$:=$' makes explicit the asymmetric nature of these equations. In other words, they are not to be rearranged to solve for their inputs. To transform these relationships from mathematical relationships to causal relations, the Causal Markov Condition is imposed, which simply assumes that arrows (and their entailed conditional independencies) represent causal dependencies \cite[p.105-6]{Peters2017}. 

The ultimate benefit of the graphical and structural model frameworks is that they, at least in principle and under some strong assumptions, enable us to use observational data to answer scientific questions such as `how?', `why?', and `what if?' \cite{Pearl2018TBOW}.

\subsection{Graphical Models}

For background on graphical models, see work by Koller and Friedman (2009) \cite{Koller2009}. We follow a similar formalism to Peters et al. (2017) \cite{Peters2017} and Strobl (2018) \cite{Strobl2018}. A graph $\mathcal{G}(\mathbf{X}, \mathcal{E})$ represents a joint distribution $P_{\mathbf{X}}$ as a factorization of $d$ variables $\mathbf{X} = \{X_1, ..., X_d\}$ using $d$ corresponding \textit{nodes/vertices} $v \in \mathbf{V}$ and connecting edges $(i, j) \in \mathcal{E}$, where $(i, j)$ indicates an edge between $v_i$ and $v_j$. If two vertices $i$ and $j$ are connected by an edge we call them \textit{adjacent}, and, can also denote this in terms of the corresponding variables $\mathbf{X}$ as $X_i \rightarrow X_j$ or $X_i \leftarrow X_j$ (directed), $X_i$ --- $X_j$ (undirected), $X_i \leftrightarrow X_j$ (bidirected), $X_i \LineCirc X_j$ or $X_i \CircLine X_j$ (partially undirected), $X_i \CircRightarrow X_j$ or $X_i \CircLeftarrow X_j$ (partially directed), or $X_i \CircCirc X_j$ (nondirected). A graph comprising entirely undirected edges forms a \textit{skeleton}. It is also possible to have self-loops, although these occur relatively infrequently in the structure discovery literature. These different edge types allow us to define a range of graph types and relationships.

An \textit{undirected path} exists if there are edges connecting two vertices regardless of the edge types between them. In contrast, a \textit{directed path} constitutes directed edges with consistent arrowhead directions. We can define a \textit{parent} $pa_j$ as a vertex $v_i$ with \textit{child} $v_j$ connected by a directed edge $X_i \rightarrow X_j$ such that $(i,j) \in \mathcal{E}$ but $(j,i) \notin \mathcal{E}$. Further upstream parents are \textit{ancestors} of downstream \textit{descendants} if there exists a directed path constituting $i_{k} \rightarrow j_{k+1}$ for all $k$ in a sequence of vertices. An \textit{immorality} or \textit{v-structure} describes when two non-adjacent vertices are parents of a common child. A \textit{collider} is a vertex where incoming directed arrows converge.

It is possible for \textit{directed cycles} to occur when following a directed path results in the visitation of a vertex more than once (\textit{e.g.}, $X_i \rightarrow X_j \rightarrow X_k \rightarrow X_i$). Many phenomena in nature exhibit cyclic properties and feedback, and ignoring this possibility has the potential to induce bias \cite{Strobl2018, Sachs2005, Ghassami2020}. If all edges are directed, and there are no cycles, we have the well-known class of \textit{Directed Acyclic Graphs} (DAGs). On the other hand, if all edges are directed but there is no restriction preventing cycles, we have a \textit{Directed Graph} (DG). 

\subsection{The Markov Assumption, $d$-Separation, and $d$-Faithfulness}
The graphs are usually assumed to fulfil the Markov property, such that the implied joint distribution factorizes according to the following recursive decomposition, characteristic of Bayesian networks \cite{Pearl2009}:
\vspace{-2mm}
\begin{equation}
    P(\mathbf{X}) = \prod_i^d P(X_i |pa_i) 
\end{equation}

This decomposition relates to the notion of $d$-separation. Two vertices $X_i$ and $X_k$ are $d$-separated by the set of vertices $\mathbf{S}$ if $X_j \in \mathbf{S}$ in any of the following structural scenarios \cite{Peters2017}:
\vspace{-1mm}
\begin{equation}
    \begin{split}
        X_i \rightarrow X_j \rightarrow X_k \\
        X_i \leftarrow X_j \leftarrow X_k \\
        X_i \leftarrow X_j \rightarrow X_k \\
    \end{split}
    \label{eq:dsep1}
\end{equation}

They are also $d$-separated if $X_j$ and none of the descendants of $X_j$ are in set $\mathbf{S}$ in the following structural scenario (collider):
\vspace{-2mm}
\begin{equation}
    \begin{split}
        X_i \rightarrow X_j \leftarrow X_k \\
    \end{split}
    \label{eq:dsep2}
\end{equation}

If the DAG's $d$-separation properties hold (an assumpion of faithfulness - see below), they imply Markovian conditional independencies in the joint distribution, which can be denoted as $X_i \indep_{P_{\mathbf{X}}} X_k | X_j$. In terms of the DAG, disjoint (\textit{i.e.}, non-overlapping) sets of variables $\mathbf{A}$ and $\mathbf{B}$ are $d$-separated by disjoint set of variables $\mathbf{S}$ in graph $\mathcal{G}$ if $\mathbf{A} \indep_{d-sep} \mathbf{B} | \mathbf{S}$ \cite{Peters2017}, and are, conversely $d$-connected if this conditional independence in the graph does not hold.

The assumption of $d$-faithfulness is that any conditional independencies implied by the graph, according to its $d$-separation properties, are reflected in the joint distribution $P_{\mathbf{X}}$. More formally \cite[p.107]{Peters2017}, for joint distribution $P_{\mathbf{X}}$ and DAG $\mathcal{G}$, the assumption of $d$-faithfulness holds if $\mathbf{A} \indep_{P_{\mathbf{X}}} \mathbf{B} | \mathbf{C} \implies  \mathbf{A} \indep_{d-sep} \mathbf{B} | \mathbf{C}$. One example of a violation of $d$-faithfulness occurs when the influence of two paths cancel each other out, resulting in a DAG with different implied conditional independencies to those present in the joint distribution.

\subsection{Markov Equivalence Class (MEC) and Completed Partially Directed Acyclic Graphs (CPDAGs)}
The conditional independence constraints implied by a graph's $d$-separation properties are not always enough to uniquely identify it. Whether a graph can be uniquely identified is known as the problem of \textit{identifiability}, and a significant body of work has been devoted to identifying scenarios for which the true graph is identifiable (\textit{e.g.}, linear functional form with non-Gaussian errors \cite{Hoyer2008}, or nonlinear functional forms with additive noise \cite{Hoyer2008b}).\footnote{One may define an SEM defined on a DAG as identifiable if there are no other SEMs that induce the same joint distribution with a different DAG \cite{Ng2020b}.} As such, there are situations in which multiple graphs satisfy the same conditional independencies. For example, conditional independence implied by $X_i \indep X_k | X_j$ is present in the graph $X_i \rightarrow X_j \rightarrow X_k$ as well as the graphs $X_i \leftarrow X_j \leftarrow X_k$ and $X_i \leftarrow X_j \rightarrow X_k$, in spite of the fact that these graphs have drastically different causal implications. The class of graphs which represent the same set of conditional independencies together constitute the \textit{Markov Equivalence Class} (MEC). Graphs belong to the same equivalence class when they have the same skeleton and the same immoralities \cite{Verma1991}. 

\begin{figure}[t!]
\centering
\includegraphics[width=0.8\linewidth]{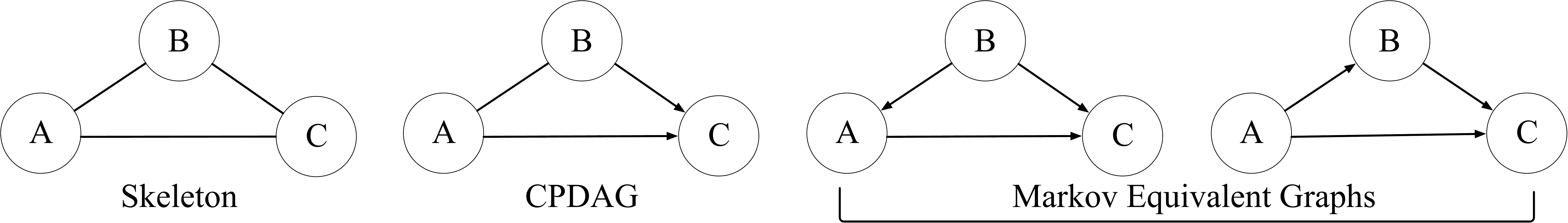}
\caption{Showing a skeleton, a CPDAG, and the Markov Equivalence set of graphs. Variable C is a collider, and so the direction of incoming arrows can be identified from conditional independencies.}
\label{fig:cpdag}
\end{figure}

\textit{Completed Partially Directed Acyclic Graphs} (CPDAGs) can be used to represent an MEC. In CPDAGs, an edge is only directed if there is only one graph in the MEC with an edge in that direction, otherwise, if there is uncertainty about the direction, it is left `non-directed' using $\CircCirc$. One might wonder whether there are any MECs without undirected edges, and indeed there are. A collider or v-structure forms an MEC with only one valid DAG: $X_i \rightarrow X_j \leftarrow X_k$. This is because conditioning on $X_j$ makes $X_i$ and $X_k$ $d$-connected. An example of a skeleton graph, a CPDAG, and corresponding MEC graphs are shown in Figure \ref{fig:cpdag}.

\subsection{Assumption: Sufficiency}
One of the challenges with using observational data is the assumption that all relevant data have been collected/observed. This is less problematic in the case of Randomized Controlled Trials (RCTs) because the randomization itself helps mitigate the effect of confounding which would otherwise imbalance the treatment and control groups.\footnote{In reality, limited sample sizes (which are often encountered with expensive RCTs) can still render this issue problematic \cite{Deaton2018}.} In observational settings, unobserved confounding can significantly bias effect estimates (even reversing their direction). Whilst it is possible to try to infer hidden confounders from observational data using latent variable models (see \textit{e.g.} \cite{Wang2019b, Louizos2017b, Vowels2020b}, a large number of causal discovery methods assume \textit{sufficiency}, which is the assumption that there are no unobserved confounders. The assumption of sufficiency is strong and may often be inappropriate or overly restrictive. If the assumption does not hold, the set of observed variables is (causally) \textit{insufficient} \cite{Bhattacharya2020} and a DAG comprising only the observed variables can not be used (and the DAG is said to not be closed under marginalization) \cite{Hu2020}.

\subsection{Acyclic Directed Mixed Graphs (AGMGs) and Maximal Ancestral Graphs (MAGs) and $m$-separation}
 In the presence of unobserved confounding, an \textit{Acyclic Directed Mixed Graph} (ADMG) may be used. ADMGs represent hidden confounding as bidirected edges. For example, the confounding relationship given by $X_i \leftarrow H \rightarrow X_j \rightarrow X_k$ can, in the absence of $H$, be represented in an ADMG as $X_i \leftrightarrow X_j \rightarrow X_k$.

\textit{Maximal Ancestral Graph} (MAG) can also be used to represent hidden confounding, and have the further capacity of representing selection bias (\textit{i.e.} as might occur when a certain sub-population is sampled). MAGs satisfy the following three properties \cite{ Richardson2002, Richardson2003, Ali2009}: (1) there are no directed cycles (acyclicity); (2) if an edge $X_i \leftrightarrow X_j$ exists (which implies $X_i$ is the \textit{spouse} of $X_j$) then there are no directed paths between $X_i$ and $X_j$; (3) if an edge $X_i$ --- $X_j$ exists (which implies $X_i$ is the \textit{neighbour} $X_j$) then $X_i$ and $X_j$ have no spouses or parents. This edge is used to represent selection bias (\textit{i.e.} where a subpopulation has been sampled according to some condition). 

The definitions of ancestor and descendent translate naturally from DAGs (see above) to MAGs, as does the definition for $d$-separation, which becomes $m$\textit{-separation}. In the latter case, the conditions for $d$-separation in Equations \ref{eq:dsep1} and \ref{eq:dsep2} hold, substituting any confounding variable relationships (\textit{e.g.}, $X_i \leftarrow H \rightarrow X_j$) with a bidirected arrow (\textit{e.g.}, $X_i \leftrightarrow  X_j$). In the presence of selection bias, a $X_i$ --- $X_j$ edge can be used. Readers are directed to \cite{ Richardson2002, Richardson2003, Ali2009} for a more detailed and formal exposition. 

\begin{figure}[t!]
\centering
\includegraphics[width=0.65\linewidth]{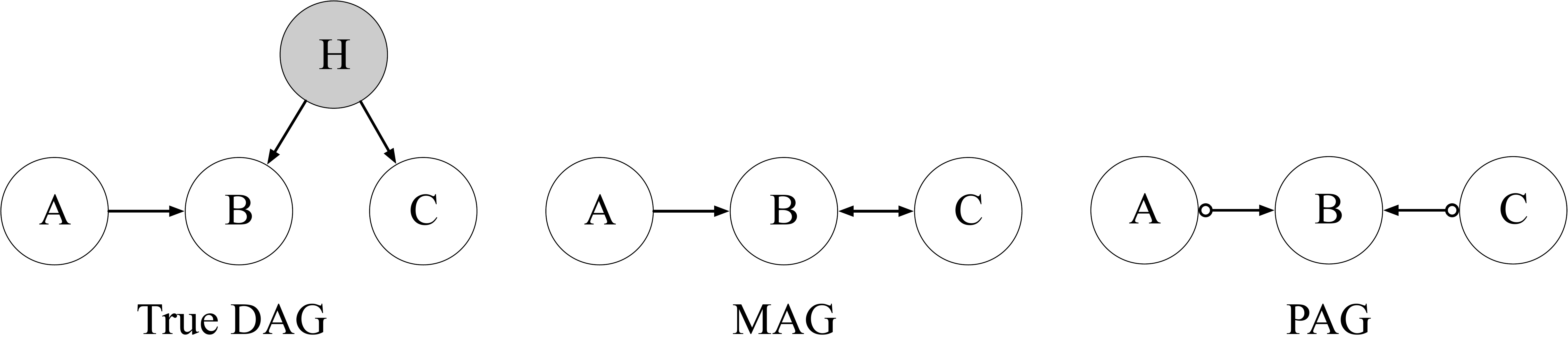}
\caption{Showing the relationship between the true DAG and its representation using a MAG and a PAG. Shaded vertex is a hidden/unobserved confounding variable. Adapted from \cite[p.179]{Peters2017}.}
\label{fig:magdagpag}
\end{figure}

The assumption of $m$-faithfulness translates naturally from $d$-faithfulness for DAGs (see above) to MAGs, according to the conditional independencies implied by $m$-separation.

\subsection{Partial Ancestral Graphs (PAGs)}
Similarly to how the MEC of a set of DAGs was represented using a CPDAG, the MEC for a set of MAGs can be representing using a \textit{Partially Ancestral Graph} (PAG). PAGs make use of edges $X_i \CircCirc X_j$, $X_i \CircRightarrow X_j$, and $X_i \LineCirc X_j$. Edges with arrowheads indicate that arrowheads are present in \textit{all} MAGs in the associated MEC. A tail (\textit{i.e.}, an edge without either a circle mark or an arrowhead) indicates that the tail is present in all MAGs in the associated MEC. Circle marks (as with CPDAGs) indicate uncertainty in the edge mark, such that the MEC contains MAGs in which the edge mark is either a tail or an arrowhead. \cite{Heinze2018, Peters2017}. An example of a DAG and its equivalent MAG and PAG are shown in Figure \ref{fig:magdagpag}.

\subsection{Other Definitions and Assumptions}
Other types of graph used to represent causal structure include Partially Oriented Induced Path Graphs (POIPGs) \cite{Spirtes2000,Peters2017}, Single World Intervention Graphs (SWIGs) \cite{Breskin2018, Richardson, Richardson2013}, $\sigma$-connection graphs \cite{Forre2018}, undirected graphs \cite{battaglia}, interaction and component graphs for dynamic systems \cite{Cummins2019}, Maximal Almost Ancestral Graphs (MAAGs) \cite{Strobl2018}, psi-ECs \cite{Jaber2020}, Patterns \cite{Zhang2020verma}, and arid, bow-free, and ancestral ADMGs \cite{Bhattacharya2020}. There are also other types of assumptions relating to the functional form of the structural relationships (\textit{e.g.}, linear or non-linear) as well as the parametric form of the marginals and the errors (\textit{e.g.}, Gaussian or non-Gaussian). In the interests of brevity, we have not discussed these additional graph-types and assumptions here, but encourage interested readers to consult the listed references.

\section{Structure Discovery Methods}
\label{sec:discovery}
We consider four approaches to structure discovery: constraint-based, score-based, those exploiting structural asymmetries, and those exploiting various forms of intervention.\footnote{There are also hybrid approaches which incorporate some combination of these classes, but we do not treat these separately.} We begin by introducing these four approaches. Each structure discovery method may be sub-categorized into those which seek to identify a graphical structure via combinatoric/search-based approaches, or those which seek to identify a graphical structure via continuous optimization. Previous reviews exist for the former (\textit{e.g.}, \cite{Heinze2018, Glymour2019, Spirtes2016}), so we primarily focus on the latter. Finally, methods may be categorized as \textit{local}, whereby edges are tested one at a time, or \textit{global}, whereby an entire graph candidate is tested.

\subsection{Constraint-Based and Score-Based Approaches}
Most constraint-based approaches test for conditional independencies in the empirical joint distribution in order to construct a graph that reflects these conditional independencies.\footnote{Other constraints exist, such as Verma constraints \cite{Verma1991, Zhang2020verma}.} According to the discussion above, there are often multiple graphs that fulfil a given set of conditional independencies, and so it is common for constraint-based approaches to output a graph representing some MEC (\textit{e.g.}, a PAG). Unfortunately, conditional independence tests require large sample sizes to be reliable, and Shah and Peters (2020) \cite{Shah2020} discuss further challenges to controlling Type I errors.\footnote{Examples of flexible conditional independence testing include GAN-based \cite{Bellot2019, Shi2020} and Kernel based \cite{Fukumizu2008, Zhang2012} methods.}

Score-based approaches test the validity of a candidate graph $\mathcal{G}$ according to some scoring function $S$. The goal is therefore stated as \cite{Peters2017}:
\vspace{-1mm}
\begin{equation}
    \hat{\mathcal{G}} = \mbox{argmax}_{\mathcal{G} \mbox{ over } \mathbf{X}} S(\mathcal{D}, \mathcal{G})
\end{equation}

where $\mathcal{D}$ represents the empirical data for variables $\mathbf{X}$. Common scoring functions include the Bayesian Information Criterion (BIC) \cite{Geiger}, the Minimum Description Length (as an approximation of Kolmogorov Complexity) \cite{Janzing2010, Grunwald2008, Kalainathan2020}, the Bayesian Gaussian equivalent (BGe) score \cite{Geiger}, the Bayesian Dirichlet equivalence (BDe) score \cite{Heckerman1995}, the Bayesian Dirichlet equivalence uniform (BDeu) score \cite{Heckerman1995}, and others \cite{Imoto2002, Hyvarinen2013, Huang2018}.

\subsection{Exploiting Structural Asymmetries}
There is no way to rule out scenarios whereby a joint distribution admits SCMs indicating either of the structural directions $X_i \rightarrow X_j$ or $X_i \leftarrow X_j$, thereby making the induction of causal directionality from observation alone, impossible. However, if some additional assumptions are made about the functional and/or parametric forms of the underlying true data-generating structure, then one can exploit asymmetries in order to identify the direction of a structural relationship. These asymmetries manifest in various ways, including non-independent errors, measures of complexity, and dependencies between marginals and cumulative distribution functions. Methods which exploit such asymmetries are typically \textit{local} methods, as they are only able to test edges one at a time (pairwise/bivariate causal directionality), or to test triples (with the third variable being an unobserved confounder) \cite{Hoyer2008}. They may, of course, be extended to construct full-graphs by iteratively testing pairwise relationships (see \textit{e.g.} the Information-Geometric Causal Inference algorithm \cite{Janzing2012}). We now briefly provide some examples of structural asymmetries, and direct interested readers to Mooij et al. (2016) \cite{Mooij2016} for a detailed review.

\subsubsection{Additive Noise}
\begin{figure}[t!]
\centering
\includegraphics[width=1\linewidth]{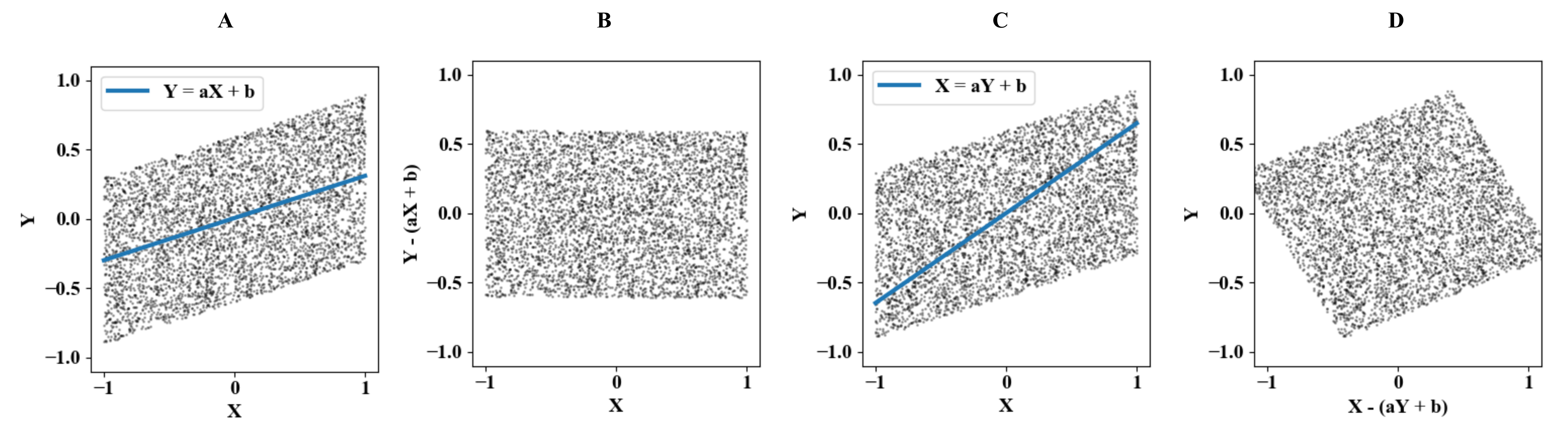}
\caption{The true structural relationship is $Y = X + U_Y$ and $X = U_X$ where $U_X$ and $U_Y$ are uniform noise sources. (A) shows the regression line when regressing $Y$ onto $X$, and (B) shows the corresponding residuals plotted against $X$. (C) shows the regression line when regressing $X$ onto $Y$ and (D) shows the corresponding residuals plotted against $Y$. Together, these demonstrate that under the assumption of linear functional form and non-Gaussian noise, the true structural direction is identifiable as the one for which $X$ is independent of the residuals, as indicated in (B). Example adapted from \cite{Peters2017}.}
\label{fig:LinNonGaussian}
\end{figure}

Given the linear structural equations $X = U_X$ and $Y = X + U_Y$ such that $U_Y \indep X$, we expect the residuals from a regression on the data from this generative model to reflect the $U_Y \indep X$ property. Interestingly, if at most $X$ or $U_Y$ is non-Gaussian, then the causal direction (\textit{i.e.} $X \rightarrow Y$) is identifiable \cite{Peters2017, HyvarinenICA2001, Kagan1973, Glymour2019}. This is illustrated in Figure \ref{fig:LinNonGaussian}. The true structural relationship $X = U_X$ and $Y = X + U_Y$ is used to generate data, where $U_X$ and $U_Y$ are non-Gaussian (they are uniformly distributed). In plot \textbf{A}, $Y$ is regressed onto $X$ (aligning with the true structural directionality), and it can be seen from plot \textbf{B} that the residuals following this regression are uncorrelated with $X$. Conversely, and as shown in plot \textbf{C}, when $X$ is regressed onto $Y$(conflicting with the true structural directionality), it can be seen in plot \textbf{D} that this results in dependence between the residuals and $Y$. 

The example given in Figure \ref{fig:LinNonGaussian} depicts the non-Gaussian, linear case. Unfortunately, the assumption that (a) the data generating process is linear and (b) that the noise are sufficiently non-Gaussian to facilitate reliable identifiability may be overly restrictive in practice. The post-non-linear additive noise model assumes that the data are generated according to the structural equations $X = U_X$ and $Y = f(X) + U_Y$ where $f$ is sufficiently non-linear, and does not assume that either $U_Y$ or $X$ are non-Gaussian. Similarly to the linear non-Gaussian case above, the post-non-linear additive noise model exhibits structural asymmetries that are reflected in the (in)dependence of regression residuals \cite{Peters2017, Glymour2019, Hoyer2008b}.

\subsubsection{Information Geometric Properties}
From a causal perspective, the information geometric approach to identifying structural directionality takes inspiration from the concept of \textit{independent mechanisms}. Assuming that the true structural direction is $X \rightarrow Y$, the concept of independent mechanisms holds that $P(X)$ contains no information about $P(Y|X)$, and vice versa. A common illustrative example \cite{Peters2017} involves measurements of temperature $Y$ at weather stations of different altitudes $X$. Regardless of the distribution of weather station altitudes $P(X)$, the mechanisms linking altitude to temperature (\textit{e.g.} the law determining the relationship between the temperature and pressure of a gas) exist independently, and changing the temperature around a weather station does not increase its altitude. 

Numerically, this scenario may be easily demonstrated by considering the inverse transform sampling method for transforming a uniform distribution $P(X)$ into a target distribution $P(Y)$ using the inverse cumulative distribution function. The uniform distribution is clearly independent of the function being used to transform it, but this independence does not hold for the transformed distribution. More generally, if $Y = f(X)$, the independence of mechanisms implies with high likelihood that $P_X$ will be independent of the mechanism $f$. The corollary is that there exists dependence between $P_Y$ and $f^{-1}$ \cite{Janzing2012}. Assuming a structural direction $X\rightarrow Y$ via function $f$, the inverse function $f^{-1}$ satisfies $\mbox{cov}[\log f^{-1}, p_Y] \geq 0$ \cite{Janzing2012, Janzing2015, Peters2017}.

It is worth noting various limitations to this approach, particularly with respect to its application to causal discovery in real-world systems. Firstly, it assumes that the mechanism $f$ is deterministic. Secondly, it assumes that $f$ is sufficiently non-linear that it may be used to identify dependence. Thirdly, real-world systems may (in addition to having non-deterministic mechanisms) demonstrate adaptation between cause and effect, such that $P_X$ is no longer independent of $f$.

\begin{figure}[t!]
\centering
\includegraphics[width=0.6\linewidth]{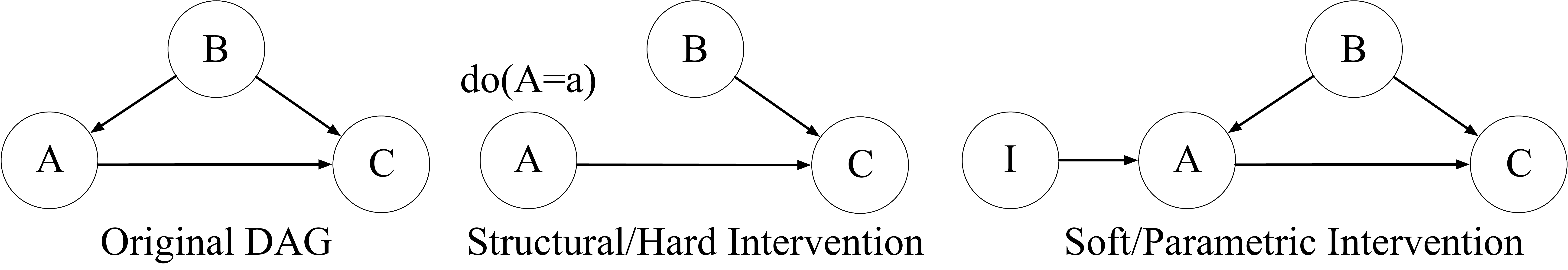}
\caption{Showing the differences between a hard/structural intervention (middle) and a soft/parametric intervention (right) on the original DAG (left). It can been that the parametric intervention preserves structural relationships. Adapted from \cite[p.986]{Eberhardt2006}.}
\label{fig:intervention}
\end{figure}

\begin{figure}[t!]
\centering
\includegraphics[width=0.55\linewidth]{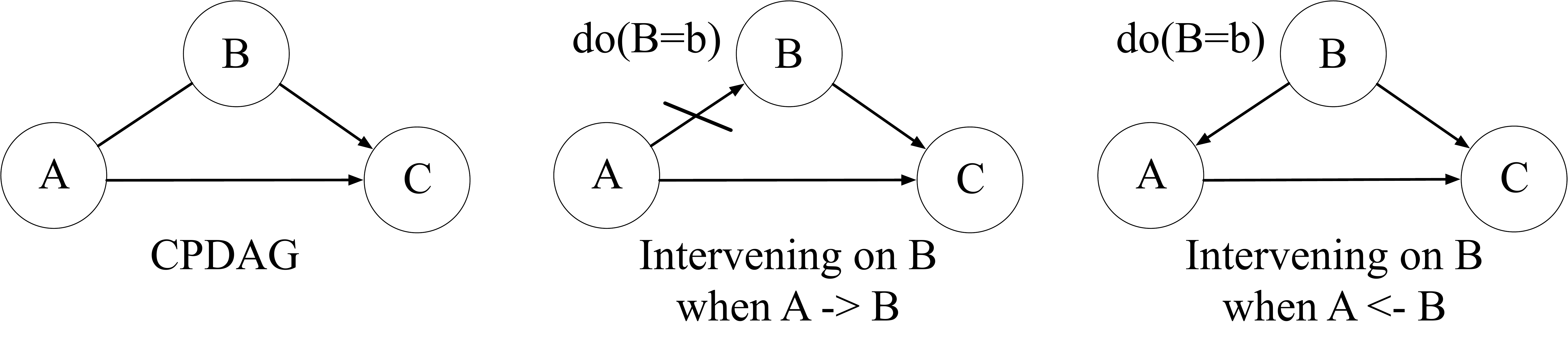}
\caption{Starting with the CPDAG on the left, where the structural direction between vertices A and B is unknown, intervening on B allows us to orient this edge. In the middle, the edge from A to B is removed following intervention on B (this is illustrated with the slash). On the right, an intervention on B does not remove the edge from B to A, and the effect of the intervention flows to A.}
\label{fig:intervention2}
\end{figure}

\subsection{Interventions and Adjustment Sets}
If interventional data are available, we are able to reduce the number of graphs in our MEC. An intervention can be denoted using Pearl's \textit{do} operator \cite{Pearl2009} such that, "for each realization of $x$ of $X$, $P(y|do(x))$ gives the probability of $Y=y$ induced by deleting from the model [$x_i = f_i(pa_i,u_i), i=1...,n,$] all equations corresponding to variables in $X$ and substituting $X=x$ in the remaining equations." Such interventions can be hard/perfect/structural/atomic/deterministic, or soft/imperfect/parametric, depending on whether a variable is set to a specific value, or whether the variable and its relationship to its neighbours is modified in some way (\textit{e.g.}, by changing the noise distribution $u$). Graphically, a hard intervention can be represented by removing all incoming arrows (from parents) to a vertex, and setting that vertex to the value $x$ \cite[p.88-91]{Peters2017}. For a structural equation model $X = U_X$, $Y=f(X) + U_Y$ and $Z = g(X) + h(Y) + U_Z$, an intervention $Y=4$ would entail $X = U_X$ (unmodified), $Y=4$ (modified), and $Z = g(X) + h(4) + U_Z$ (modified). Thus it can be seen that only $Y$ and its descendants have been affected by the intervention, leaving $X$ unchanged. In contrast to hard interventions, a parametric intervention preserves the structure of the intervention itself, introducing an additional vertex and affecting the conditional distribution of the intervened variable. Parametric interventions also preserve any correlations deriving from unobserved/hidden confounders \cite{Eberhardt2006}. This difference is illustrated in Figure \ref{fig:intervention}.

In order to demonstrate how interventions can be used to narrow the equivalence set (and in some cases make the true graph identifiable), consider the graphs in Figure \ref{fig:intervention2}. Starting with the CPDAG on the left, where the edge from A to C is undirected because the direction of the edge cannot be ascertained from conditional independencies alone. Intervening (hard) on B allows us to orient this edge by comparing the resulting distribution under intervention. If the edge is oriented $A \rightarrow B$ then the intervention has the effect of `removing' this edge. Conversely, if the edge is oriented $B \rightarrow A$ then the intervention does nothing to remove this arrow, and the downstream variable $A$ should change accordingly.

Another way to view interventions from the perspective of independencies is to consider their formulation in terms of \textit{adjustment sets}. Following \cite{PetersSID2015, Cinelli2020, Pearl2009}: $p(y|do(X=x)) = p(y)$ if  $Y$ is a parent of $X$ (\textit{i.e.}, $Y \in pa_X$). In words, intervening on $X$ does not change $y$ because $X$ is `downstream' of $Y$. Secondly, if $Y \not \in pa_X$, then:
\vspace{-1mm}
\begin{equation}
    p(y|do(X=x)) = \sum_{pa_X}p(y|x, pa_X)p(pa_X)
\end{equation}

Here, the marginalized interventional distribution $p(y|do(X=x))$ is being computed using the \textit{adjustment} set (in this case $pa_X$ is a valid adjustment set). More generally, the interventional distribution can be calculated as:
\vspace{-1mm}
\begin{equation}
    p(y|do(X=x)) = \sum_\mathbf{z} p(y|x, \mathbf{z})p(\mathbf{z})
\end{equation}

when $\mathbf{z}$ is a valid adjustment set for this particular interventional distribution. Note that sets including mediators and descendants of mediators (where $B$ in the graph $A \rightarrow B \rightarrow C$ is a mediator) are \textit{not} valid adjustment sets for finding $C | do(A)$. See Cinelli et al. (2020) \cite{Cinelli2020} for a "Crash course in good and bad controls". 

For a detailed review of different types of interventions and their implications, readers are directed to Eberhardt \& Scheines (2006) \cite{Eberhardt2006}. Suffice to say there are many ways to leverage different types of intervention, including multiple interventions on different vertices, or single interventions applied to multiple nodes. Finally, there is work investigating the use of data representing unknown or uncertain interventions, whereby it is not known which variables have been intervened on \cite{Ke2020b,  Rothenhausler2015, Eaton2007, Mooij2016b}. The use of intervention also yields what is known as an Interventional Equivalence Class (IEC), representing the set of graphs compatible with a given intervention(s).

\subsection{Causality Over Time}
 Consider a graph $X \rightarrow Y$ for the case where $X$ and $Y$ vary over time. In this scenario, a single right-arrow is not sufficient to detail whether $X$ causes $Y$ on an intra-timepoint\footnote{It is generally accepted that an effect has to follow the cause in time, thereby precluding contemporaneous effects. However, in cases where the sampling rate is too low to capture this delay, it is reasonable to model the effects as instantaneous.} basis (\textit{i.e}, contemporaneously), or on an inter-timepoint (\textit{i.e.}, lagged) basis. Indeed, at different points in time, and over different lags, the direction of causality may switch. In these cases it is common to \textit{unroll} the graph over time, such that each instance of variables $X^t$ and $Y^t$ and their structural relationships over time are modelled explicitly.
 
 There are then two types of causality considered in the context of time series. The most common (which is generally considered to be the industry standard, particularly in economics) is \textit{Granger} causality \cite{Granger1980}. If a variable $X$ `Granger-causes' $Y$,  then it means that $Y^t \not\!\perp\!\!\!\perp X^{<t} | Y^{<t}$ \cite[pp.207]{Peters2017}, where $<t$ indicates timepoints previous to $t$. If this is the case, then the predictability of $Y^t$ will decrease when $X^{<t}$ is removed from the model (because $X$ contains unique information for predicting $Y$), when accounting for previous values of $Y$.
 
 The first thing to note about Granger causality is that it tends to fail in the presence of contemporaneous effects \cite[pp.207]{Peters2017}, owing to difficulties with identifiability. The second, and perhaps more important aspect of Granger causality, is that it is only applicable if \textit{separability} holds. Separability refers to the independence of the variables in the absence of causal interactions. Unfortunately, this is rarely the case in dynamic systems, where the current state of a variable may be heavily determined by the past of another (\textit{e.g.}, consider a predator-prey model, where both population levels are always functions of each other). 
 
 This failure of Granger causality was noted by Granger himself, and has motivated the development and application of \textit{dynamic-causality}; in particular, methods deriving from Sugihara et al.`s Convergent Cross Mapping methods \cite{Sugihara2012, Ye2015}. The methods operate using \textit{time delayed embeddings} or \textit{shadow manifolds}. These shadow manifolds are constructed by concatenating time-lagged versions of the original time series. This process has been shown according to Takens' theorem \cite{Takens1981}, to be sufficient in recovering the dynamics of the full system even if only one, or a limited number, of observational variables are used. Dynamic or CCM-causality has shown great promise in applications to ecosystems and genetics \cite{Sugihara2012, Ye2015}. The general idea behind these methods is to exploit asymmetries that exist between the compactness of neighbourhoods of points in the shadow manifolds. If $X \rightarrow  Y$ in a CCM-causal sense, then points which are tightly clustered in the `effect shadow manifold' of $Y$ should also be tightly clustered in the `cause shadow manifold' of $X$. This characteristic does not hold in the reverse direction (nor if there is no causal interaction in either direction), and this asymmetry enables us to identify causal interactions and directionality.

\subsection{Evaluation Metrics}
There are a number of common metrics used for evaluating the performance of causal discovery algorithms. The metrics given below are those used to evaluate the success of edge discovery. Other score metrics can be used to measure model fit (such as the log likelihood or the Bayesian Information Criterion). For a more detailed discussion on structural discovery metrics, readers are directed to work by de Jongh (2009) \cite{Jongh2009}.

\textbf{True Positive Rate (TPR)} \cite[p.383]{Heinze2018}:
Assuming the probability of an edge $p(a_{ij})$ can be thresholded by $t \in (0,1) $, TPR is defined as $\mbox{TPR}_t = |\{(i,j):p(a_{ij})\geq t\} \cap \mathbf{S}| / |\mathbf{S}|$ where $\mathbf{S}$ is the set of ground truth edges (\textit{i.e.}, $\{(i,j):a^*_{ij} = 1\}$). 

\textbf{False Positive Rate (FPR)} \cite[p.383]{Heinze2018}: Assuming the probability of an edge $p(a_{ij})$ can be thresholded by $t \in (0,1) $, FPR is defined as $\mbox{FPR}_t = |\{(i,j):p(a_{ij})\geq t\} \cap \hat{\mathbf{S}}| / |\hat{\mathbf{S}}|$ where $\hat{\mathbf{S}}$ is the set of ground truth missing edges (\textit{i.e.}, $\{(i,j):a^*_{ij} = 0\}$).

\textbf{Area Over Curve (AOC)} \cite[p.383]{Heinze2018}: Simply $(1-\mbox{Area Under Curve (AUC)}$ for the AUC of $(FPR_t, TPR_t)$ where the threshold $t$ is varied between 0 and 1. Either the AUC or AOC can be used as a structure discovery performance metric.

\textbf{Structural Hamming Distance (SHD)}: Is the number of required changes to the graph for it to match the ground truth. It is the sum of missing edges, extra edges, and incorrect edges \cite{Jongh2009}.

\textbf{Structural Interventional Distance (SID)} \cite{PetersSID2015}: Is a count of the number of vertex pairs $(i,j)$ for which the intervention $p(x_j | do(X_i = x))$ would be incorrect if the estimated graph (as opposed to the ground-truth graph) were used for the associated adjustment set. It is therefore particularly well suited for causal inference tasks \cite[p.7]{Lachapelle2020}.

\section{Combinatoric/Search Based Approaches}
\label{sec:comb}
The number of possible DAGs increases super-exponentially with the number of variables \cite{Robinson1973}. As noted by Peters et al. (2017) \cite{Peters2017}, the number of possible DAGs for 10 variables is $> 4\times10^{18}$. As such, the search problem is NP-hard \cite{Chickering1996}, and this will later motivate the use of continuous optimization approaches to graph learning.

\begin{table}[hb!]
\centering
\fontsize{5.6}{6.6}\selectfont
\begin{tabular}{llllllll}
\toprule
\textbf{Method}  & \textbf{Year} & \textbf{Type} & \textbf{Suff.} & \textbf{Faith.} & \textbf{Acycl.} & \textbf{Interv.} & \textbf{Output} \\ \midrule
PC 	\cite{Spirtes2000} &	1993 &	constraint	& yes	& yes &	yes &	no &	CPDAG \\
CCD	 \cite{Richardson1996}	 & 1996	 & constraint &	yes	 & yes	& both	& no &	PAG \\
FCI 	\cite{Spirtes2000}	 & 2000	 & constraint & 	no & 	yes & 	yes	 & no & 	PAG \\
TPDA  	\cite{Cheng2002} & 	2002	 & constraint & 	yes & 	yes	 & yes & 	no	 & CPDAG \\
CPC	  \cite{Ramsey2006} & 	2006	 & constraint & 	yes & 	relaxed	 & yes & 	no & 	CPDAG \\
KCL   	\cite{Sun2007}	 & 2007	 & constraint & 	yes	 & yes	 & yes & 	no & 	CPDAG \\
ION	  \cite{Tillman2008} & 	2008 & 	constraint	 & no & 	yes	 & yes	 & no & 	PAG \\
IDA  	\cite{Maathuis2009}	 & 2009 & 	constraint & 	yes	 & yes & 	yes	 & yes & 	DAG \\
cSAT+  	\cite{Triantafillou2010} & 	2010 & 	constraint & 	no & 	yes & 	yes & 	no & 	PCG \\
KCI-test \cite{Zhang2012} & 2012 & constraint & yes	& yes &	yes &	no &	CPDAG \\
RFCI  	\cite{Colombo2012RFCI}	 & 2012 & 	constraint & 	no	 & yes & 	yes & 	no	 & PAG \\
CHC	  \cite{Gamez2012} & 	2012 & 	constraint & 	yes & 	yes & 	yes & 	no & 	PDAG \\
SAT 	\cite{Hyttinen2013}	 & 2013 & 	constraint & 	no	 & yes & 	no & 	yes & 	DG \\
Parallel-PC	  \cite{Le2014} & 	2014 & 	constraint & 	yes	 & yes & 	yes	 & no & 	CPDAG \\
RPC   	\cite{Harris2013} & 	2013 & 	constraint	 & yes & 	yes	 & yes & 	no	 & CPDAG \\
PC-stable  	\cite{Colombo2014} & 	2014 & 	constraint	 & both	 & yes & 	both & 	no	 & CPDAG \\
COmbINE	  \cite{Triantafillou2015}	 & 2015	 & constraint & 	no & 	yes & 	yes & 	yes	 & summary SMCMs \\
backshift  	\cite{Rothenhausler2015} & 	2015 & 	- & 	no	 & no & 	no & 	yes	 & DG \\
IGSP	\cite{Yang2018} &	2018	 &  constraint & yes & relaxed & yes & yes & I-MEC \\	
$\sigma$-CG  	\cite{Forre2018} & 	2018 & 	constraint & 	no	 & yes & 	no & 	yes & 	$\sigma$-connection graphs \\
CCI	  \cite{Strobl2018} & 	2018 & 	constraint	 & no & 	yes	 & no & 	no & 	MAAG \\
FCI-soft  	\cite{Kocaoglu2019} & 	2019 & 	constraint & 	no	 & relaxed	 & yes & 	yes	 & I-MEC \\
IBSSI  	\cite{Chicharro2020} & 	2020 & 	constraint	 & no & 	yes	 & yes	 & yes	 & DAG \\
CD-NOD	 \cite{Huang2020}	 & 2020	 & constraint & 	no & 	yes & 	both & 	yes	 & ----- \\
psi-FCI	  \cite{Jaber2020}	 & 2020	 & constraint & 	no & 	relaxed	 & yes	 & yes & 	Psi-EC \\
LCDI	  \cite{Zhang2020verma}	 & 2020	 & constraint & 	no & 	yes	 & yes	 & yes & Pattern \\
EG  	\cite{Elidan2009} & 	2009 & 	score & 	yes	 & yes & 	yes	 & no & 	BT-DAG \\
TWILP  	\cite{Parviainen2014} & 	2014 & 	score & 	yes	 & yes	 & yes & 	no	 & BT-DAG \\
K2 	\cite{Cooper1992} & 	1992	 & score & 	no	 & yes & 	yes	 & no & 	CPDAG \\
LB-MDL  	\cite{Lam1994}	 & 1994	 & score & 	yes	 & yes & 	yes	 & no & 	DAG \\
HGC	  \cite{Heckerman1995} & 	1995 & 	score & 	yes	 & yes & 	yes & 	no & 	CPDAG \\
GES	  \cite{Chickering2002} & 	2002 & 	score & 	yes	 & yes	 & yes	 & no & 	CPDAG \\
OS	  \cite{Teyssier2005} & 	2005 & 	score & 	yes	 & yes	 & yes & 	no & 	DAG \\
HGL	  \cite{He2005} & 	2005 & 	score	 & yes & 	yes	 & yes & 	yes	 & CPDAG \\
Meinshausen  	\cite{Meinshausen2006} & 	2006 & 	score & yes	 & 	- & 	no & 	no & 	UG \\
Graphical Lasso	  \cite{Friedman2008} & 	2008 & 	score	 & yes & 	- & 	no & 	no & 	UG \\
BC  	\cite{Banerjee2008}	 & 2008	 & score & 	yes	 & - & 	no & 	no & 	UG \\
TC  	\cite{Pellet2008} & 	2008 & 	score & yes	 & 	yes	 & yes & 	no & 	CPDAG \\
HG	  \cite{He2008} & 	2008 & 	score & 	yes	 & yes & 	yes & 	yes	 & DAG \\
Adaptive Lasso  	\cite{Shojaie2010} & 	2010 & 	score & 	yes	 & yes & 	yes & 	no & 	DAG \\
GIES  	\cite{Hauser2012} & 	2012 & 	score	 & yes & 	yes	 & yes	 & yes & 	PDAG \\
CD  	\cite{Fu2013} & 	2013 & 	score	 & yes & 	yes & 	yes  & 	yes	 & DAG \\
GBN learner  	\cite{Geer2013}	 & 2013 & 	score & 	yes & 	no & 	yes & 	no	 & CPDAG \\
GES-mod	  \cite{Barba2013} & 	2013 & 	score & 	yes & 	yes	 & yes & 	no & 	CPDAG \\
Pen-PC	  \cite{Ha2015} & 	2015	 & score & 	yes	 & yes	 & yes & 	no & 	CPDAG \\
Scalable GBN  	\cite{Aragam2015} & 	2015 & 	score & 	yes	 & no & 	yes & 	no	 & DAG \\
K-A*   	\cite{Scanagatta2016} & 	2016 & 	score	 & yes	 & yes & 	yes & 	no & 	DAG \\
NS-DIST	  \cite{Han2016} & 	2016 & 	score	 & yes & 	no & 	yes & 	yes	 & DAG \\
MIP-GD	  \cite{Park2017}	 & 2017 & 	score & 	yes & 	yes & 	yes	 & no & 	CPDAG \\
CD2	  \cite{Gu2018} & 	2018 & 	score	 & yes & 	yes & 	yes & 	yes	 & DAG \\
SP  	\cite{Raskutti2018}	 & 2018	 & score & 	yes	 & relaxed & 	yes	 & no & 	CPDAG \\
VAR	  \cite{Young2018}	 & 2018	 & score & 	yes	 & yes & 	both & 	no & 	DG \\
GSF   	\cite{Huang2018} & 	2018 & 	score & 	yes	 & yes & 	yes & 	no & 	CPDAG \\
bQCD  	\cite{Tagasovska2020} & 	2020 & 	score	 & yes	 & yes & 	yes	 & no & 	Bi \\
GCL 	\cite{Varando2020} & 	2020 & 	score & 	no & 	- & 	no	 & no & 	GCLM \\
GGIM  	\cite{Fitch2020} & 	2020 & 	score	 & yes & 	no & 	no & 	no & 	GGIM  \\
GYKZ  	\cite{Ghassami2020} & 	2020 & 	score & 	yes	 & yes & 	both & 	no	 & DG \\
SLARAC etc.	  \cite{Weichwald2020} & 	2020 & 	score	 & Granger	 & - & 	- & 	no & 	Bi \\
Order-MCMC	  \cite{Friedman2003} & 	2003 & 	sampling	 & yes & 	yes	 & yes & 	no & 	DAG \\
OG	  \cite{Ellis2008}	 & 2008	 & sampling	 & yes & 	yes & 	yes	 & yes	 & DAG \\
EE-DAG	  \cite{Zhou2011}	 & 2011 & 	sampling & 	yes	 & yes & 	yes	 & yes & 	DAG \\
ZIPBN  	\cite{Choi2020} & 	2020 & 	sampling & 	yes	 & no	 & yes & 	no	 & DAG \\
LiNGAM	  \cite{Shimizu2006LiNGAM}	 & 2006	 & asymmetries & 	yes	 & no & 	yes & 	no & 	DAG \\
LV LiNGAM  	\cite{Hoyer2008} & 	2008	 & asymmetries & 	no & 	yes	 & yes & 	no & 	DAG \\
non-linear ANM  	\cite{Hoyer2008b} & 	2008 & 	asymmetries & 	yes	 & yes & 	yes	 & no & 	DAG \\
CAN	  \cite{janzing2009} & 	2009 & 	asymmetries	 & no & 	yes & 	yes & 	no & 	Bi/tri \\
CCM \cite{Sugihara2012} & 2012 & asymmetries & - & - & no & no &Bi\\
IGCI	\cite{Janzing2012} & 	2012 & 	asymmetries	 & yes & 	yes & 	yes	 & no & 	Bi \\
KCDC	\cite{Mitrovic2018}	 & 2018	 & asymmetries & 	yes & 	yes	 & yes & 	no & 	Bi \\
MMHC	\cite{Tsamardinos2006MMHC} & 	2006 & 	hybrid	 & yes	 & yes	 & yes & 	no	 & DAG \\
ARGES	\cite{Nandy2018} & 	2018 & 	hybrid & 	yes	 & yes	 & yes & 	no & 	CPDAG \\ \bottomrule
\end{tabular}
\caption{This table comprises a list of non-continuous optimization based approaches to causal discovery (\textit{i.e.}, combinatoric, search-based, SAT-solver). Provides indication of assumptions of sufficiency (`Suff.'), faithfulness (`Faith.'), acyclicity (`Acycl.'), as well as whether the method leverages forms of intervention (`Interv.'). `Bi' indicates bivariate cause-effect pairs (possibly multivariate). N.B. If the output is `DAG' this does not necessarily imply that the method identifies the \textit{true} DAG.}
\label{tab:combtable}
\end{table}

Table \ref{tab:combtable} presents a non-exhaustive list of methods which do not use continuous optimization. In other words, they include primarily combinatoric/search-based approaches to structure discovery. The table presents the type of approach used: constraint-based, score-based, asymmetry-based, hybrid, and sampling-based (which measure belief in a proposed graph structure by sampling from a posterior). In addition, the table provides the associated assumptions: Sufficiency (\textit{i.e.}, whether it assumes there are no hidden variables), Faithfulness (some methods achieve a less severe/relaxed form of faithfulness); and Acyclicity (some methods can learn feedback loops and cycles). Finally, the table indicates whether the method leverages interventions, and indicates the method's output (CPDAG, PAG, etc.).

\section{Continuous Optimization Based Approaches}
\label{sec:cont}
The primary focus of this survey is to review continuous optimization based methods for structure discovery. Continuous optimizaion methods are pervasive in the field of deep learning, whereby highly parameterized networks are optimized using variations on gradient descent \cite{goodfellow}. Increased compute (particularly with the advent of GPUs) make the task of learning from large, high-dimensional datasets feasible. Recently, there have been an increasing number of methods which seek to learn structure from data, whilst leveraging the advantages of continuous optimization. This has resulted in the confluence of black-box deep learning approaches, and structure discovery. These continuous optimization approaches recast the combinatoric graph-search problem into a continuous optimization problem (specifically, an Equality Constrained Program) \cite{Zheng2018b}. In Equation~\ref{eq:notears}, the left hand side represents the traditional approach, which seeks the adjacency matrix $\mathbf{A}$ that minimizes some score function $S(\mathbf{A})$, subject to the implied $d$-vertex graph $\mathcal{G}(A)$ being in the set of valid DAGs. The right hand side represents a characterization of the continuous optimization problem which, again, seeks the adjacency matrix $\mathbf{A}$ that minimizes some score function $S(\mathbf{A})$, but this time subject to the constraint $h(\mathbf{A}) = 0 $. Here, $h$ is the function used to enforce acyclicity in the inferred graph. 
\vspace{-2mm}
\begin{equation}
\begin{array}{cl}
\min _{\mathbf{A} \in \mathbb{R}^{d \times d}} S(\mathbf{A}) & \; \; \; \; \; \; \min _{\mathbf{A} \in \mathbb{R}^{d \times d}} S(\mathbf{A}) \\ \text { subject to } \mathcal{G}(\mathbf{A}) \in \mathrm{DAGs} & \;\;\;\;\;\text { subject to } h(\mathbf{A})=0
\end{array}
\label{eq:notears}
\end{equation}

\vspace{-1mm}

The increased popularity of structure discovery in deep learning is not without sound motivation, with arguments that disentangled, structured, and symbolic representations are key to the next generation of AI, as well as robust cross-domain performance, transfer learning, and interpretability \cite{Greff2020, Bengio2019b, bengio1, Garcez2020}. Researchers have noted three primary approaches to learning representations of the world: (1) distributed, (2) structured and symbolic, and (3) a hybrid of (1) and (2). Most basic neural networks perform distributed learning and there is no clear separation of high-level semantics..

\begin{table}[hb!]
\centering
\scriptsize
\begin{tabular}{llllll}
\toprule
\textbf{Method}  & \textbf{Year} & \textbf{Data} & \textbf{Acycl.} & \textbf{Interv.} & \textbf{Output} \\ \midrule
CMS \cite{Ma2014} & 2014 & low & - & no &Bi \\
NO TEARS	\cite{Zheng2018b}&	2018 &	low &	yes &	no &	DAG \\
CGNN	\cite{Goudet2018}&	2018 & low &  yes & no & DAG \\				
Graphite	\cite{Grover2019}&	2019 &  low/medium & no & no & UG \\		SAM	\cite{Kalainathan2020}&	2019 & low/medium & yes & no & DAG \\	
DAG-GNN	\cite{Yu2019}&	2019	 & low & yes &  no & DAG \\	
GAE	\cite{Ng2019}&	2019		 & low &  yes& no & DAG\\			
NO BEARS	\cite{Lee2019NOBEARS}&	2019	 & low/medium/high & yes & no & DAG \\	
Meta-Transfer	\cite{bengio2019}&	2019	 &Bi& yes & yes &Bi\\	
DEAR	\cite{Shen2020}	&2020			 & high & yes & no & - \\		
CAN	\cite{Moraffah2020}	&2020 &	low/medium/high	& yes &	no &	DAG \\
NO FEARS	\cite{Wei2020}	&2020				 &  low & yes & no & DAG\\	
GOLEM	\cite{Ng2020}&	2020		 & low & yes & no  & DAG\\			
ABIC	\cite{Bhattacharya2020}	&2020		 & low & yes & no & ADMG/PAG \\			
DYNOTEARS	\cite{Pamfil2020}	&2020		 & low & yes & no & SVAR \\			
SDI	\cite{Ke2020b}	&2020				 & low & yes & yes & DAG \\	
AEQ	\cite{Galanti2020}&	2020			 &Bi& - & no & direction\\		
RL-BIC	\cite{Zhu2020b}&	2020		 & low & yes & no & DAG \\			
CRN	\cite{Ke2020}&	2020			 & low & yes  &  yes & DAG\\		
ACD	\cite{Lowe2020ACD}	&2020		 & low & Granger & no & time-series DAG\\			
V-CDN	\cite{Li2020b}	&2020		 & high & Granger & no  & time-series DAG\\			
CASTLE (reg.)	\cite{Kyono2020}	&2020		 & low/medium & yes & no & DAG\\			
GranDAG	\cite{Lachapelle2020}	&2020		 & low &yes & no  & DAG\\			
MaskedNN	\cite{Ng2020b}	&2020	 & low & yes & no & DAG\\	    			
CausalVAE	\cite{Yang2020}	&2020			 & high & yes & yes & DAG\\		
CAREFL	\cite{Khemakhem2020}&	2020	 & low & yes & no & DAG / Bi\\				
Varando	\cite{Varando2020}&	2020		 & low & yes& no& DAG \\			
NO TEARS+	\cite{Zheng2020}&	2020	 & low& yes& no& DAG \\				
ICL	\cite{Wang2020}	&2020				 & low  & yes & no & DAG\\	
LEAST	\cite{Zhu2020LEAST}	&2020 & low/medium/high & yes & no & DAG 	\\ \bottomrule
\end{tabular}
\caption{This table comprises a list of continuous optimization based approaches to causal discovery. `Data' indicates the dimensionality of the data the method has been demonstrated to handle. `Bi' indicates bivariate cause-effect pairs (possibly multivariate), `low' indicates <100 vertices, `medium' indicates >100, and `high' indicates either dimensionality >10,000 or data which are not already projected into a causal/semantic space (\textit{e.g.}, image data). `Acycl.' indicates whether the method enforces acyclicity, and `Interv.' indicates the use of interventions during learning. Please consult the main test for further details of each method. N.B. If the output is `DAG' this is not meant to necessarily imply the method identifies the \textit{true} DAG.}
\label{tab:conttable}
\end{table}

Conversely, DAGs are highly structured and facilitate causal reasoning. However, such reasoning is only possible if one already has access to variables which represent high-level semantic concepts, which is not the case when learning from raw video data, for example. Hence the motivation for hybrids which can be used to `learn' or infer high-level representations as well as the structured relations between them. Examples of hybrid approaches include methods such as Recurrent Independent Mechanisms \cite{Goyal2019}, graph networks \cite{battaglia, Scarselli2009,  Beanini2020}, and a large body of work on scene understanding \cite{Mao2019,Nash2017,Yi2020, Nanbo2020, Jiang2020}. The debate as to how much structural inductive bias / constraint is required for an algorithm to reason effectively is ongoing \cite{Ding2020, battaglia}. Indeed, finding a DAG to represent complex phenomena (such as natural language) is non-trivial and potentially impossible.

In this section we review the recent evolution of continuous optimization based approaches to structure learning, and Table \ref{tab:conttable} presents a non-exhaustive list of such methods.

\subsection{Cross Map Smoothness (CMS, 2014)}
The CMS algorithm \cite{Ma2014} is intended to identify causal directionality between time varying variables in dynamic systems. The method is inspired by convergent cross mapping methods \cite{Sugihara2012, Ye2015} which operate using \textit{time delayed embeddings} or \textit{shadow manifolds}. In order to ascertain whether two variables $X$ and $Y$ from a time varying dynamical system are causally related, CMS uses a radial basis function neural network to map between the shadow embeddings for $X$ and $Y$. The asymmetry in the error when mappying from $X$ to $Y$, compared with the error when mapping from $Y$ to $X$, is used as a proxy to infer causal directionality. Note that this method is only demonstrated to work when the input variables are univariate, but may be used multiple times to ascertain the causal structure of more than two observational variables.

\subsection{DAGs with NO TEARS (2018)}
The recent (2018) method DAGs with NO TEARS (Non-combinatoric Optimization via Trace Exponential Augmented lagRangian Structure learning) \cite{Zheng2018b} is generally considered as the first to recast the combinatoric graph search problem as a continuous optimization problem (see Equation \ref{eq:notears}). The function $h$ for enforcing acyclicity is derived to be:
\vspace{-1mm}
\begin{equation}
    h(\mathbf{A}) = tr(e^{\mathbf{A}\odot \mathbf{A}}) - d = 0
    \label{eq:notearsacyclicity}
\end{equation}
\vspace{-5mm}

In practice, $h(A)$ may be small but non-zero, and edges may require some thresholding. One of the disadvantages of this acyclicity constraint is that the matrix exponential requires $\mathcal{O}(d^3)$ computations, and subsequent methods seek to improve on this. The structural model learnt is linear such that $X_j = a_j^T \mathbf{X} + U_j$, where $a_j$ is the weight in the adjacency matrix corresponding with the edges into $X_j$, (the noise variables are not assumed to be Gaussian). NO TEARS uses a least-squares loss with an $l1$ penalty to encourage sparsity, and their objective is optimized using the Augmented Lagrangian method \cite{Nemirovski1999} with L-BFGS \cite{Byrd1995}. As well as synthetic data, NO TEARS is also evaluated on the proteins and phospholipid dataset by Sachs et al. (2005) \cite{Sachs2005}. Despite the fact that the formulated optimization problem does not guarantee an optimal solution, their results demonstrate close-to-optimal results on the chosen datasets.

\subsection{Causal Generative Neural Network (CGNN, 2018)}
CGNN \cite{Goudet2018} combines graph learning with continuous optimization, neural networks, and hill-climbing or Tabu search. The neural networks are used to learn the functions mapping variables (\textit{e.g.} see the SEM breakdown in Figure \ref{fig:sem}), where the variables themselves are selected according to the output of a greedy-search  algorithm. The motivation for the neural network is that they do not impose restrictions on the functional form \textit{a priori} and therefore "let the data speak" \cite{vanderLaan2011}. The networks are trained using the Adam \cite{adam} optimizer with a Maximum Mean Discrepancy (MMD) \cite{Gretton2007} score function. During training, the edges are directed in order to minimize this discrepancy, and following training, the graph is adjusted to remove cycles. CGNN incorporates a hill-climbing search algorithm to optimize the structure of the DAG, and then the network optimization resumes. This training cycle is repeated to convergence, and each edge has an associated score representing its contribution to the global fit. They use a thresholding function to regularize the number of edges in the graph. Finally, their method includes a means to identify possible hidden confounding, by leveraging the fact that confounding can be modelled as correlations/associations between the (otherwise) exogenous latent random variables.

\subsection{Graphite (2019)}
Graphite \cite{Grover2019} is a generative neural network model incorporating a graph neural network \cite{Scarselli2009} encoder, where latent variables are inferred using black-box variational inference \cite{Ranganath2013, rezende2, kingma, blei2}. Graphite takes in graphical/network data, and infers a posterior latent distribution over these data. The network is trained to reconstruct the graph which is parameterized using a symmetric, weighted adjacency matrix (\textit{i.e.} the graph is undirected). The method is shown to perform well on data with as many as 19,717 vertices.

\subsection{Structural Agnostic Modeling (SAM, 2019)}
SAM \cite{Kalainathan2019, Kalainathan2020} is a neural network approach that is intended to address the limitations of CGNN. These limitations are CGNN's quadratic complexity (due to the calculation of the MMD), and the scalability issues that arise due to CGNNs use of a greedy-search. SAM addresses these two limitations with the use of adversarial training \cite{goodfellow2}, and by making the mechanism which optimizes the DAG part of end-to-end training. Their score function is a log-likelihood loss with two model complexity regularizers: One that penalizes the model, on a per-vertex basis, by an amount proportional to the number of vertex parents; and one which acts as neural network parameter/weight decay. It uses an acyclicity constraint which is similar to the one in NO TEARS \cite{Zheng2018b} to encourage DAG-ness:
\vspace{-2mm}
\begin{equation}
    \sum_{k=1}^{d}\frac{tr(A_j)}{k!} = 0
\end{equation}

Here, $A$ is what they call a structural gate, which performs the same function as an adjacency matrix. The neural network parameterization of the structural equation model is $X_j = L_{j,H+1}\circ \sigma \circ ...L_{j,1}([\mathbf{a}_j \odot \mathbf{X}, U_j]$. In words, the stack of $H$ neural network layers $L_H$ and non-linearities $\sigma$ for each variable $j$ is used as the function over the Hadamard product between the data $\mathbf{X}$ and a binary vector form of the adjacency matrix $A$, s.t. $a_{i,j} = 1$ iff there is an edge $X_i \rightarrow X_j$. They provide a detailed theoretical analysis of their method, showing how the global training objective constitues a combination of a structural component (which seeks the CPDAG) and a functional component (which exploits asymmetries). They assume both faithfulness and sufficiency, and evaluate on a range of low to medium dimensionality datasets (the highest number of dimensions is approximately 6000 (DREAM5 \cite{Marbach2012DREAM}).

\subsection{DAG Graph Neural Network (DAG-GNN, 2019)}
DAG-GNN \cite{Yu2019} extends NO TEARS by incorporating neural network functions $f$ and black-box variational inference such that the score function is the Evidence Lower BOund (ELBO) \cite{Ranganath2013, rezende2, kingma, blei2}. The method assumes faithfulness, and infers a latent posterior $\mathbf{Z}$: 
\vspace{-1mm}
\begin{equation}
    \mathbf{Z} = f_4((\mathbf{I}-\mathbf{A}^T)f_3(\mathbf{X})
    \label{eq:daggnn1}
\end{equation}

where $A$ is a weighted adjacency matrix, and $\mathbf{X}$ may comprise vector-valued variables. DAG-GNN recovers the observations with a decoder:
\vspace{-1mm}
\begin{equation}
\mathbf{X} = f_2((\mathbf{I} - \mathbf{A}^T)^{-1}f_1(\mathbf{Z}))
\label{eq:daggnn2}
\end{equation}

Together, Equations \ref{eq:daggnn1} and \ref{eq:daggnn2} constitute a variational autoencoder \cite{kingma, rezende2}. Noting that if $f_2$ is invertible, then:
\vspace{-1mm}
\begin{equation}
    f_2^{-1}(\mathbf{X}) = \mathbf{A}^Tf_2^{-1}(\mathbf{X}) + f_1(\mathbf{Z})
\end{equation}
which is a generalization of the linear SEM model $\mathbf{X} = \mathbf{A}^T\mathbf{X}+\mathbf{Z}$. Acyclicity is enforced using a constraint derived from the one employed in NO TEARS \cite{Zheng2018b} as:
\vspace{-1mm}
\begin{equation}
    tr[(\mathbf{I} + \alpha \mathbf{A} \odot \mathbf{A})^d] - d = 0
\end{equation}
where $\alpha$ acts as a hyperparameter on this constraint. This formulation of the acyclicity constraint is justified on the basis of that it is preferred over a calculation that involves the matrix exponential (as appears in Equation \ref{eq:notearsacyclicity}). Similarly to NO TEARS, they also use the augmented Lagriangian approach to optimization. They evaluate on low-dimensional data such as the proteins and phospholipid dataset by Sachs et al. (2005) \cite{Sachs2005}.

\subsection{Graph AutoEncoder (GAE, 2019)}

GAE \cite{Ng2019} extends NO TEARS and DAG-GNN formulations for structure learning into a graph autoencoder model, facilitating non-linear structural relationships and vector-valued variables. They model structure in the same way as DAG-GNN, and draw a connection to graph convolutional neural networks \cite{Kipf2017}:
\vspace{-1mm}
\begin{equation}
    f(X_j, \mathbf{A}) = f_2(\mathbf{A}^T f_1(X_j))
\end{equation}
where $f_1$ and $f_2$ are multilayer perceptrons (MLPs). Similarly to NO TEARS, and DAG-GNN, they also use the augmented Lagrangian method with Adam \cite{adam} for constrained optimization. Their acyclicity constraint is identical to the one used in NO TEARS (Equation \ref{eq:notearsacyclicity}). They demonstrate that GAE performs significantly better than NO TEARS and DAG-GNN, particularly as the number of vertices in the graph increases, and also highlight that training time is much shorter.

\subsection{Meta-Transfer Objectives (2019)}
Bengio et al. \cite{bengio2019} identify that if the correct causal direction is known, then learning algorithms adapt faster under distributional shift (\textit{i.e.} intervention), than they do if it makes the incorrect assumptions about the direction. This is demonstrated by comparing the adaptation rates, and formulating a meta-learning objective that accounts for the rates of learning under different directional assumptions.

\subsection{Scaling Structural Learning with NO BEARS (2019)}

The NO BEARS method \cite{Lee2019NOBEARS} is partly motivated by the computational complexity $\mathcal{O}(d^3)$ associated with the matrix exponential in the NO TEARS acyclicity constraint. They reformulate the acyclicity constraint by using the spectral radius of a matrix. Normally the spectral radius also requires $\mathcal{O}(d^3)$ operations, but they present an approximation that takes only $\mathcal{O}(d^2)$. The spectral radius is the maximum magnitude of eigenvalues, and the authors show how it forms an upper bound on the original NO TEARS acyclicity constraint. Rather than using neural networks to increase the flexibility of the structural functions, they use a polynomial (order 3) regression. NO BEARS is demonstrated to scale well even on data with as many as 12,800 vertices.

\subsection{Disentangled gEnerative cAusal Representation Learning (DEAR 2020)}

DEAR \cite{Shen2020} combines a Variational AutoEncoder (VAE) \cite{kingma, rezende2} with an adversarial loss \cite{goodfellow2} in order to infer a latent space with "causal" structure. Strictly, this is not a causal discovery method, because they assume the `super-graph' is given, and they learn the associated weights and parameters. The latent space is given supervision in the form of labels for the generative factors. The latent structure is defined as:
 \vspace{-1mm}
\begin{equation}
    z = f((\mathbf{I} - \mathbf{A}^T)^{-1}h(\epsilon))
\end{equation}

Here, $A$ is a weighted adjacency matrix, $f$ and $h$ are neural networks, and $\epsilon$ is noise sampled from a prior distribution. DEAR is notable for its use of high-dimensional data with semantic labels. It maps from image data to the structured latent space, where the labels provide a form of supervision.

\subsection{Causal Adversarial Network (CAN 2020)}

CAN \cite{Moraffah2020} is a Generative Adversarial Network (GAN) \cite{goodfellow2} that facilitates interventional sampling from a structural graph (which the authors refer to as a causal graph) at inference time. It comprises a Label Generation Network, which learns a graph from the dataset labels, and a Conditional Image Generation Network, which generates the images conditioned on the interventional distribution specified by the user at inference time. Their generator is a function of an adjacency matrix applied to the noise vectors as $\mathbf{X} = G((\mathbf{I} - \mathbf{A}^T)^{-1}\mathbf{Z})$ where $X$ is a sample from the join distribution, $\mathbf{Z}$ is random noise, $A$ is a weighted adjacency matrix, $\mathbf{I}$ is the identity matrix, and $G$ is the non-linear generator function. In order to impose acyclicity they leverage an equality constraint \cite{Yu2019}, such that acyclicity occurs if:
\vspace{-1mm}
\begin{equation}
    tr[(\mathbf{I} + \beta \mathbf{A} \odot \mathbf{A})^d] - d = 0
\end{equation}

where  $d$ is the number of vertices in the graph, `$tr$' is the trace operator, $\odot$ is the Hadamard product, and $\beta$ is a non-zero hyperparameter.

As well as evaluating CAN on CelebA \cite{celebA} image data (including the generation of interventional samples), they also evaluate it on the more traditional CHILD \cite{Spiegelhalter1992} and Alarm \cite{Beinlich1989} datasets showing performance competitive with state-of-the-art.

\subsection{DAGs with NO FEARS}
NO FEARS \cite{Wei2020} revisits and updates aspects of DAG-GNN and NO TEARS. They provide a detailed analysis of the acyclicity constraint of NO TEARS (see Equation \ref{eq:notearsacyclicity}) and show that, following the augmented Lagrangian optimization, it is not guaranteed to converge to a feasible solution of the intend constraint (\textit{i.e.}, when $h(\mathbf{A}) = 0$). Instead of a constraint that depends on $\mathbf{A}\odot \mathbf{A}$, they propose one that depends only on the absolute value $|\mathbf{A}|$, on the basis that there is a connection with the $l1$ penalty and sparsity. Following some modifications to make the absolute value function differentiable, the authors modify existing algorithms with knowledge derived through theoretic analysis, and show their proposal to improve all baselines (including combinatoric approaches).

\subsection{Gradient-based Optimization of dag-penalized Likelihood for learning linEar dag Models (GOLEM, 2020)}
Following in a similar vain to other works such as NO BEARS, DAG-GNN, and NO FEARS, GOLEM \cite{Ng2020} examines the acyclicity constraint of NO FEARS. They also note that NO TEARS uses a least-squares score function, and improve on this by proposing a score function that directly maximizes the data likelihood. The authors show that in the linear Gaussian case and under mild assumptions (such as faithfulness), a likelihood-based objective with `soft' sparsity regularization is sufficient to asymptotically identify a quasi-equivalent (see original paper for definition) DAG and that a hard acyclicity constraint is not required. Further, in the linear non-Gaussian scenario, they explain how an acyclicity constraint is not needed in the asymptotic regime, although it may be necessary with finite samples. Finally, they explain how it is sufficient to have a `soft' acyclicity penalty, instead of a hard constraint, which greatly reduces the complexity of the optimization problem. They propose their own objective, including a likelihood based score with an $l1$ regularizer and soft acyclicity constraint, which they optimize using Adam \cite{adam}. Some post-processing is undertaken to threshold edges in order to guarantee acyclicity. The primary distinctions from NO TEARS are, therefore, (a) the likelihood based score function, and (b) the use of a soft (rather than hard) aycyclicity penalty.

\subsection{Approximate Bayesian Information Criterion for Differentiable Causal Discovery Under Unmeasured Confounding (ABIC, 2020)}
ABIC \cite{Bhattacharya2020} extends the continuous optimization paradigm to discover various types (ancestral, arid, bow-free) of ADMGs which account for unmeasured confounding. In the linear SEM case, unmeasured confounders manifest as correlated errors, which are represented in a \textit{second} adjacency matrix. They present three differentiable constraints which can be used to discover a particular type of ADMG. They use the BIC criterion as the primary objective/score function. The parameters are optimized using a Residual Iterative Conditional Fitting algorithm \cite{Drton2009}.

\subsection{DYNOTEARS (2020)}

DYNOTEARS \cite{Pamfil2020} seeks to discover structure in time series data, which is a topic we have not covered in Section 2 of this work. By using second order optimization, DYNOTEARS seeks to learn a Structural Vector AutoRegressive (SVAR) model, which is also a form of dynamic Bayesian network. This is argued to be important on the basis that temporal dynamics are an essential part of real-world systems, which cannot be captured using a static graph model. They assume that variables potentially affect each other both contemporaneously, and in a time-lagged manner. DYNOTEARS is, therefore, not Granger causal, because it accounts for contemporaneous effects \cite[p.203-208]{Peters2017}. They model two adjacency matrices, $\mathbf{W}$ and $\mathbf{A}$, for the intra-slice and inter-slice graph edges, respectively. Because the edges represented in $\mathbf{A}$ only go forward in time, only $\mathbf{W}$ needs an acyclicity constraint. They use the same constraint as NO TEARS (see Equation \ref{eq:notearsacyclicity}, and incorporate it into an augmented Lagrangian problem which is optimized using L-BFGS-B \cite{Zhu1997}. Following optimization, and similarly to other methods using acyclicity constraints, they threshold edges with weights close to 0. DYNOTEARS is evaluated on S\&P 500 returns data with 97 vertices, and on DREAM4 with 100 vertices \cite{Marbach2009}.

\subsection{Structural Discovery from Intervention (SDI, 2020)}
SDI \cite{Ke2020b} is a neural network method that assumes faithfulness and sufficiency, and which attempts to discover structure using data which have been subject to unknown interventions. SDI is restricted to discrete, categorical variables with no missingness; it assumes the available interventions are sparse and only effect a single (possibly unknown) variable; the interventions may be soft; and there are no compounding interventions (\textit{i.e.}, only one or less interventions occur in the data). 

The method is trained in three stages which repeat until convergence. The first stage is concerned with updating the functional parameters (those which map between vertices). The procedure involves randomly drawing data samples and graph configurations, and optimizing the functional parameters using the log-likelihood as a score function. In the second stage the structural parameters are updated (those which model the edges between vertices), and interventional (unknown) data are sampled. The variable subject to intervention is predicted using a simple heuristic; namely, that the variable exhibiting the greatest reduction in log-likelihood is predicted on the basis that it is a poor fit to the observational distribution. Given a new set of interventional data and sampled graphs, these graphs can be scored whilst masking the intervened variable. In the third stage, and following \cite{bengio2019}, the REINFORCE algorithm \cite{REINFORCE} is used to update the discrete structural parameters. 

They apply an acyclicity constraint which is derived from Equation \ref{eq:notearsacyclicity} as: $\sum_{i\neq j} \mbox{cosh}(\sigma(a_{ij})\sigma(a_{ji})) $, where $a_{ij}$ is the structural parameter linking variable $i$ to $j$, and $\sigma$ is the sigmoid function. The method is evaluated on low-dimensional data ($d < 100$), and is shown to exceed state-of-the-art on a number of benchmark datasets.

\subsection{AutoEncoder Complexity (AEQ, 2020)}
The authors of the AEQ method \cite{Galanti2020} develop a score function based on autoencoder reconstruction error for discovering the directionality of vector valued cause-effect pairs. Their key result is that the SEM $Y = g(f(X), U)$ only holds in one direction if $X$ and $Y$ are vectors and $g$ and $f$ are neural network functions. They extend this result to univariate $X$ by creating multivariate versions of the variable based on a sorted concatenation of slices of the original. The complexity of this multivariate surrogate is then measured using an autoencoder reconstruction error (they use an $l2$ loss). For a cause-effect pair, the variable with the higher loss is likely to be the cause. In the case where the original variables are multivariate, they propose an adversarial conditional independence method that discriminates between joint distributions and the product of the marginals (resembling a mutual information proxy).

\subsection{Causal Discovery with Reinforcement Learning (RL-BIC, 2020)}
The authors of RL-BIC \cite{Zhu2020b} take a reinforcement learning approach to causal discovery. They generate directed graphs using an encoder-decoder neural network model, which forms the `actor'. The output of the encoder-decoder is the proposed graph, which is scored using the BIC in order to generate a reward signal. A critic is used to update the proposed graphs and therefore also to drive the optimization of the neural network parameters. They assume an additive noise model $X_i = f_i(pa_i) + U_i$ as well as faithfulness and causal sufficiency. Their output graph is represented using a binary adjacency matrix. They mask out $(i,i)$ edges to prevent self-loops, and incorporate an adapted form of the NO TEARS acyclicity penalty:
\begin{equation}
    h(\mathbf{A}) = tr(e^{\mathbf{A}})- d = 0
\end{equation}
In order to guarantee acyclicity (in the event that $h(\mathbf{A})$ is small but non-zero), they augment it with a hard indicator function penalty that acts on whether the graph is a valid DAG or not. All generated graphs are stored during training, and the one with the best score is chosen, and this graph is finally pruned to reduce false discovery. The method is trained using poly-gradient method and REINFORCE \cite{REINFORCE} with Adam \cite{adam}, and evaluated on relatively small graphs ($\leq 30$ nodes).

\subsection{Meta-Learning Neural Causal Representations (CRN, 2020)}
Ke et al. (2020) \cite{Ke2020} propose a meta-learning neural network method that leverages continuous representations of graphs and which assumes causal sufficiency. Training is split into episodes where, for each episode, a graph is proposed and used to generate data for the duration of the episode. The episode is further split into $k$ time points, and for each time point a random intervention is undertaken on the graph and data is generated. The model is then asked to predict the outcome of the intervention, and thereby ends up `learning' the causal relationships between the variables in the graph. They also propose a Causal Relational Network (CRN), which accumulates information about the interventions and graphs over time (similar to an LSTM \cite{hochreiterLSTM}). They use a graph decoder (the gradients from which are not backpropagated to the rest of the network) in order to validate the graph's continuous representation against the ground truth graph. It is shown that CRNs learn new causal models quickly and efficiently. Interestingly, there is no discussion about (a)cyclicity, but nonetheless their intention is to learn DAGs.

\subsection{Amortized Causal Discovery (ACD, 2020)}

ACD \cite{Lowe2020ACD} is a Granger-causality non-linear time-series method which leverages black-box variational inference \cite{Ranganath2013, rezende2, kingma, blei2} to infer a latent posterior graph. Granger causality assumes there are no contemporaneous effects \cite[p.203-208]{Peters2017}. The method is demonstrated to perform well under hidden confounding (and so does not assume causal sufficiency). 

ACD learns from samples with different causal relationships but shared dynamics. This is motivated using an example from neuroscience. They use the encoder to infer the causal graph from a particular sample, and a decoder which models the dynamics and takes past samples and the inferred graph in order to predict the future. Specifically, for sample $\mathbf{X}_s$ with graph encoder $f$ and decoder dynamics model $g$, the future is predicted as $\mathbf{X}_s^{t+1} = g(\mathbf{X}_s^{\leq t}, f(\mathbf{X}_s))$. The graph is inferred from the entire sample, and the dynamics model $g$ is used to predict the future given the inferred graph and a portion of the past.

\subsection{Causal Discovery from Video (V-CDN, 2020)}
The authors of V-CDN \cite{Li2020b} use unsupervised key-point detection on video data in order to build a causal representation of these points. It is a Granger-causal non-linear time-series method which leverages black-box variational inference \cite{Ranganath2013, rezende2, kingma, blei2} and deep neural networks to infer a latent graph which explains the structural relationships between the deteceted keypoints. They integrate a dynamics module to facilitate future prediction.

\subsection{Causal Structure Learning (CASTLE, 2020)}
The authors of CASTLE \cite{Kyono2020} propose causal discovery as an auxiliary task which helps to regularize a supervised predictive model. The motivation is that, by identifying key causal factors, the model avoids overfitting to potential confounders which hurt model robustness and generalizability. Specifically, a neural network model attempts to identify the DAG that explains the structural relationships between the observed variables, and this task is built into an autoencoder \cite{Kramer1991} framework. Their structural model is non-parametric, following the form $X_i = f_i(pa_i, U_i)$ and using an acyclicity constraint:
\vspace{-2mm}
\begin{equation}
    h(\mathbf{A}) = (tr(e^{\mathbf{A}\odot \mathbf{A}}) - d - 1)^2
\end{equation}
which, they explain, also forces the autoencoder to reconstruct only the input variables which have neighbours.

\subsection{Gradient Based Neural DAG Learning (GranDAG, 2020)}
In a similar vain to other methods, GranDAG \cite{Lachapelle2020} seeks to expand upon NO TEARS in order to deal with non-linear relationships through the use of neural networks. They follow the non-linear additive noise structural model of the form $X_j = f_j(pa_j) + U_j$, where each function $f_j$ is parameterized as a fully-connected neural network. In order to maintain an independence of mechanisms which corresponds with the independence implied by an adjacency matrix, they formulate \textit{neural network paths} and a  \textit{connectivity matrix}, resembling previous work by Germain et al. (2015) \cite{Germain2015MADE}. The connectivity matrix $\mathbf{C}_j$ is essentially the matrix product of all neural network weights in a single neural network (\textit{i.e.}, parameterizing one $f_j$). This product results in $\mathbf{C}_j \in \mathbb{R}^{m \times d}$ where $m$ is the number of parameters needed to specify a chosen distribution for $X_j$ (\textit{e.g.}, a Gaussian has two parameters), and $d$ is the number of variables.  If $\mathbf{C}_{j,ki}= 0$ then the input $i$ is independent of output $k$ for variable $X_j$. Note that $f_j$ takes as input $X_{-j}$ (where the variable of interest $j$ is masked to zero). The connectivity matrix is then used to define their weighted adjacency matrix, such that the adjacency matrix $\mathbf{A} \in \mathbb{R}^{d\times d}$ depends on all neural network weights from all neural networks. They define the weighted adjacency matrix and substitute it into the NO TEARS acyclicity constraint as:
\vspace{-1mm}
\begin{equation}
    h(\mathbf{A}) = tr(e^{\mathbf{A}}) - d = 0
\end{equation}

For learning they employ the augmented Lagrangian formulation, using a log-likelihood score function, and threshold the resulting edges for $h(\mathbf{A}){ij})$ close to zero. They demonstrate that their continuous optimization based approach exceeds the performance of combinatoric approaches such as PC \cite{Spirtes2000}, as well as other continuous optimization based approaches e.g. NO TEARS and DAG-GNN.

\subsection{Masked Gradient Based Structure Learning (MaskedNN, 2020)}
The researchers behind \cite{Ng2020b} attempt to improve on NO TEARS \cite{Zheng2018b} using neural networks. They assume an additive noise SEM of the form $X_j = f_j(pa_j) + U_j$, and explain how their method can be directly extended from handling scalar variables to vector valued variables. They provide a discussion on identifiability (something which a number of methods in both the combinatoric and continuous optimization literature tend to omit). They provide an overview of the gradual evolution from NO TEARS (which assumes linear SEMs), via DAG-GNN \cite{Yu2019}, GAE \cite{Ng2019} and GraNDAG \cite{Lachapelle2020} (which handle non-linear SEMs), but highlight that these methods do not provide an in depth discussion about identifiability. They also highlight that the use of non-linear transformations on the adjacency matrices in DAG-GNN and GAE may affect their causal interpretability.

MaskedNN uses a binary adjacency matrix  $\mathbf{A}$ (rather than weighted), which is integrated into their SEM as: $X_j = h_j(\mathbf{A}_j \odot \mathbf{X}) + U_j$ and refer to this as an Augmented SEM (ASEM). Their discussion on identifiability states that their method can learn a Super-graph of the true graph, and further utilize thresholding and Causal Additive Model \cite{Buhlmann2014} based pruning to remove spurious edges under mild conditions. They leverage the Gumbel-softmax trick \cite{Maddison2017, jang2017} to incorporate discrete learning (in view of the binary adjacency matrix) into an augmented Lagrangian 1st order continuous optimization based approach with an Adam optimizer \cite{adam}).

\subsection{Causal Variational AutoEncoder (CausalVAE, 2020)}

The creators of CausalVAE \cite{Yang2020} argue that whilst many disentangled representation learning methods assume independence between latent factors \cite{locatello, higgins, kumar2}, most latent factors behind real-world phenomena exhibit causal dependencies. They propose the use of a Variational AutoEncoder \cite{rezende2, kingma}. The latent space of a VAE is usually parameterized by a set of exogeneous factors (often modelled as a multivariate, isotropic Gaussian). CausalVAE integrates a \textit{Causal Layer} which transforms these exogenous latent factors into endogenous factors which reflect the causal semantics of the data. They assume a linear SEM following the form $\mathbf{Z} = \mathbf{A}^T \mathbf{Z} + \mathbf{U}$ where $\mathbf{Z}$ are the inferred latent factors following the application of the adjacency matrix $\mathbf{A}$. They integrate supervision in the form of semantic labels $\mathbf{Y}$ to condition the posterior $p(\mathbf{Z}|\mathbf{Y})$, which forces identifiability.

These factors (which now reflect semantic quantities according to the provided supervision) are then passed to a masking layer, similar to the one used in MaskedNN. They then apply $Z_j = g_j(\mathbf{A}_j \odot \mathbf{Z}) + U_j $ where $g$ are nonlinear and invertible functions. $\mathbf{A}_j \odot \mathbf{Z}$ yields a vector only containing parental information, because the adjacency matrix effectively masks non-parents. The authors explain how this masking layer facilitates interventional queries. In order to learn the causal structure, they incorporate the structural inductive prior into the supervised loss function:
\vspace{-1mm}
\begin{equation}
    l_y = \mathbb{E}_q || \mathbf{Y} - \sigma (\mathbf{A}^T\mathbf{Y})||^2_2
\end{equation}

where $q$ is the approx posterior distribution. They incorporate the NO TEARS acyclicity constraint:
\vspace{-2mm}
\begin{equation}
    h(\mathbf{A}) = tr((\mathbf{I}+ \mathbf{A}\odot \mathbf{A})^d)-d = 0
\end{equation}

The method is evaluated on the CelebA \cite{celebA} dataset, as well as a synthetic data of a pendulum casting a shadow from a light. The second dataset is used to demonstrate the interventions - they intervene (for example) on the position of the light in order to demonstrate the independence of the position of the pendulum as well as the dependence with the shadow.

\subsection{Causal AutoRegressive Flows (CAREFL, 2020)}
In CAREFL \cite{Khemakhem2020}, the authors combine causal discovery with the deep learning framework known as normalizing flows \cite{Kobyzev2020, rezende}. Normalizing flows provide a means to construct generative models which have the capacity to model complex densities using invertible transformations of a basic and tractable density. They enable the exact computation of the log-likelihood (which constitute their learning objective) via the use of the change of variables formula and inverse log Jacobian determinant. Specifically, they use autoregressive flows, which are a form of normalizing flow for which the transformations are affine and have simple, lower-triangular Jacobians \cite{Kingma2016, Marino2019}.

The authors consider an SEM in terms of a \textit{causal ordering}, whereby, according to the SEM/DAG, there exists a permutation of the vertices that corresponds to the order of specified dependencies. For example, a parent vertex precedes a child vertex in the causal ordering. The generic additive noise SEM $X_j = f_j(pa_j)+ U_j$ can be written in terms of a causal ordering $\pi$ as $X_j = f_j(\mathbf{X}_{<\pi(j)}) + U_j$ (which is assumed for CAREFL), where $X_{<\pi(j)}$ represents variables that precede $X_j$ in the causal order (including its parents). This latter form is shown to bear resemblance to the autoregressive flow model with a few constraints. The CAREFL method is shown to be flexible enough to answer both counterfactual and interventional queries. As well as outputting a DAG, the method can also be used to judge causal direction by using the log-likelihood to score different directions.

\subsection{DAGs without Imposing Acyclicity (NODAG, 2020)}
Varando (2020) \cite{Varando2020} proposes a proximal gradient \cite{Parikh2014} optimization objective that yields a linear SEM and corresponding DAG without requiring an acyclicity constraint. The method derives the novel objective by framing the learning problem in terms of sparse matrix factorization, and the resulting method NODAG is shown to be both effective and efficient.

\subsection{NO TEARS+ (2020)}
A number of the same authors from NO TEARS revisit their original work and update it. We refer to this later work as NO TEARS+ \cite{Zheng2020}, which seeks to extend NO TEARS acyclicity constraint to handle nonparametric, general models of the form $g_j(f_j(X))$ (which subsumes additive noise models, linear models, and generalized linear models). This model does not utilize an adjacency matrix, and thus they frame acyclicity in terms of partial derivatives (an idea they attribute to Rosasco et al. (2013) \cite{Rosasco2013}) such that $[\mathbf{W}(f)]_{kj} := ||\partial_k f_j||_2$. This states that the dependency structure between variable $k$ and the function $f_j$ (which is described by the DAG represented in matrix $\mathbf{W}$) is the $l2$ norm of the partial derivative of $f_j$ with respect to $X_k$. They integrate a multi-layer perceptron into their derived framework (as well as a number of other variations) and demonstrate it's effective performance and efficiency.

\subsection{Imputated Causal Learning (ICL, 2020)}

The authors of ICL \cite{Wang2020} focus on the problem of structure discovery under the missing-data setting, and provide definitions and examples of three types of missingness: Missing At Random (MAR), Missing Completely At Random (MCAR), and Missing Not At Random (MNAR). They propose the use of Generative Adversarial Networks (GANs) \cite{goodfellow2} and Variational AutoEncoders (VAEs) \cite{rezende2, kingma}. ICL takes incomplete data and simultaneously imputes the missing data using the GAN, in order to match the generated distribution to the empirical distribution. The task of the discriminator in the GAN is to differentiate between observed versus generated data. The skeleton graph is estimated using a method following DAG-GNN \cite{Yu2019}. Following this, the edges in the skeleton are oriented following a method proposed by Cai et al. (2019) \cite{Cai2019} which is based on the additive noise model for causal direction identification.

\begin{table}[h!]
\centering
\scriptsize
\begin{tabular}{ll}
\hline
\textbf{Method}  & \textbf{Keywords \& Software}  \\ \hline
causaleffect \cite{causaleffect} & general causality, R\\
daggity \cite{dagitty}&  general causality, R\\
dosearch \cite{dosearch} &causal effect identification, R\\
Causal Discovery Toolbox \cite{Kalainathan2019b} &causal discovery, Python\\
pcalg \cite{Kalisch2012} &causal discovery, R\\
bnlearn \cite{Scutari2017} &causal discovery, R\\
rEDM \cite{rEDM} & dynamic modeling and convergent cross mapping, R \\
DoWhy \cite{dowhy}  & general causality, Python\\
CausalImpact \cite{Brodersen2015} &intervention, time series, R\\
causal-cmd \cite{pycausal} & general causality, Python (py-causal) \& JAVA + CLI \\
\hline \hline
\end{tabular}
\caption{This table comprises a list of Python and R packages for general causal inference and structure discovery. CLI = command line interface}
\label{tab:packages}
\end{table}

\vspace{-1.0cm}

\subsection{Scalable Learning for Bayesian Networks (LEAST, 2020)}

The authors of LEAST \cite{Zhu2020LEAST} propose a new acyclicity constraint, which improves upon the $\mathcal{O}(d^3)$ cost of NO TEARS \cite{Zheng2018b}. To do this, they first consider:
\begin{equation}
    h(\mathbf{S}) = tr(e^\mathbf{S}) - d = 0
\end{equation}

to be the NO TEARS constraint, where $\mathbf{S} = \mathbf{A}\odot \mathbf{A}$. This was subsequently altered by \cite{Yu2019} to:
\vspace{-1mm}
\begin{equation}
    g(\mathbf{S}) = tr((\mathbf{I} + \mathbf{S})^d) - d = tr(\sum_{k=1}^d\frac{d}{k}\mathbf{S}^k) =0 
\end{equation}

\vspace{-3mm}
 on the basis that $e^\mathbf{S} = \sum_{i=0}^{\inf} \frac{\mathbf{S}^k}{k!}$, where $k$ is the length of a cycle. The authors of LEAST argue that both of these have drawbacks relating to $\mathcal{O}(d^3)$ complexity, as well as storage of  $e^S$. They note that NO BEARS \cite{Lee2019NOBEARS} framed the problem in terms of a spectral radius (the absolute value of the largest Eigenvalue of $S$). However, this also requires $\mathcal{O}(d^3)$ computation, so they derive an upper bound $\bar{\delta}$ on this spectral radius as:
 \vspace{-1mm}
 \begin{equation}
 \begin{split}
     \bar{\delta}^{(k)} = \sum_{i=1}^db^{(k)}[i] \; \; \; \mbox{where} \\
     b^{(k)} = (r(\mathbf{S}^{(k)}))^\alpha \odot (c(\mathbf{S}^{(k)}))^{1-\alpha} \; \; \; \mbox{and} \\
     \mathbf{S}^{(k+1)} = (D^{(k)})^{-1}\mathbf{S}^{(k)}D^{(k)} \;\;\;\; \mbox{and} \\
     D^{(k)} = \mbox{Diag}(b^{(k)})
     \end{split}
 \end{equation}
\vspace{-1mm}

Combining a computable form for this upper bound with the least squares objective and $l1$ regularization, they show that this new objective is nearer to $\mathcal{O}(d)$, and trains between 5 and 15 times faster than NO TEARS. Note that edge thresholding is still required. They demonstrate the benefits of this speedup by evaluating on both small graphs, as well as graphs with as many as 160,000 vertices.


\begin{table}[hb!]
\centering
\scriptsize
\begin{tabular}{lll}
\toprule
\textbf{Dataset}  & \textbf{Vertices} & \textbf{Notes}  \\ \midrule
Multi-body Interaction \cite{Li2020b} &  - & up to 5 moving balls with physical interactions/relations \\
Fabric deformation \cite{Li2020b} &  - &applying forces to different fabrics \\
Cause-effect pairs \cite{Tagasovska2020} & 2 & bivariate distributions\\
Cause-effect pairs \cite{Mooij2010b} & 2 & bivariate distributions \\
Cause-effect pairs (Tuebingen) \cite{Mooij2016, Mooij2014} & 2 & bivariate  distributions \\
SynTReN \cite{Syntren} & user specified & synthetic gene expression data\\
Sachs \cite{Sachs2005} &  11 & proteins and phospholipids in human cells \\
Scale-Free Graphs \cite{Barabasi1999} & user specified & preferential attachment graph generation law \\
Erdos-R\'{e}nyi Graphs (\textit{e.g.} \cite{Lachapelle2020}) & user specified & adds edges with probability $p=\frac{2e}{d^2-d}$ \\
Linear, GP Add, GP Mix, Sigmoid Add and Sigmoid Mix & - & mixed graph data \\
CausalWorld \cite{Ahmed2020} & - & comprehensive robotics dataset\\
MPI3D \cite{Gondal2019} & - & visual disentanglement dataset \\
Pendulum-light-shadow \cite{Yang2020} & - & image data \\
Phase coupled oscillator \cite{Kuramoto1975} & - & physical relations \\
NetSim \cite{Smith2011} & user specified & fMRI data simulation\\
Temperature \cite{Lowe2020ACD} & & \\
BnLearn \cite{Scutari2017}  & -  & Repository \\
DREAM series \cite{Marbach2009, Marbach2012DREAM} & up to 6000 & simulated and in-vivo gene regulation networks \\
Causality 4 Climate \cite{Runge2019} & - & climate change time series competition data \\
Archaeology \cite{Huang2018} & 8 & archaeology data \\ 
S\&P500 & 500 & time series / stock returns\\
\bottomrule
\end{tabular}
\caption{This table comprises a list of datasets that have been used for testing structure discovery methods.}
\label{tab:data}
\end{table}

\vspace{-1cm}
\section{Summary and Discussion}
\label{sec:summary}
We have attempted to present the relevant background, definitions, assumptions, approaches to causal discovery, common evaluation metrics, as well as providing a brief review of combinatoric methods, and a detailed review of continuous optimization based methods. In terms of additional resources, a range of software packages exist for undertaking causal inference and structure discovery and we have provided a list in Table \ref{tab:packages} for convenience. Also, in Table \ref{tab:data} we provide a list of datasets used for causal discovery. Note that not all of these datasets are readily available.
 Finally, we encourage readers to explore various additional references and commentaries. These include: A discussion of the relevance of causality to machine learning \cite{Scholkopf2019, Neto2020, Gong2020}; Commentaries on the nature of causality \cite{LewisCausation1973, MenziesCausation2020}; alternative reviews on causal inference and causal discovery \cite{Guo2020, Yao2020, Glymour2019, Heinze2018, Spirtes2016}; reviews with a focus on time-series causal inference and discovery \cite{Eichler2013}; frameworks for dynamical SCMs with ODEs \cite{Peters2020, Mooij2013}; guides on the foundations for causal discovery \cite{Eberhardt2017}; some example applications \cite{Sani2020, Lin2020c, LopezPaz2016, anonICLR, Zhu2020b}; textbooks on causal inference and causal discovery \cite{Peters2017, Spirtes2000, Pearl2009}.

\subsection{Opportunities and Future Directions}

One of the main advantages to combinatoric approaches to structure discovery relates to the provision of guarantees for identifying the true graph, or at least the true equivalence class. This advantage comes at a significant cost, however, because such approaches are limited to low-dimensional problems (or low-cardinality graphs) due to the super-exponential search space. One might expect, then, that even though the continuous optimization approaches are confronted with a non-trivial, non-convex solution space, they might at least scale to larger problems. Unfortunately, and as can be seen from Table \ref{tab:conttable}, most continuous optimization approaches have only been evaluated on low-dimensional problems. This seems to be due to the fact that the most common acyclicity constraint, namely the one in Equation \ref{eq:notearsacyclicity} from NO TEARS \cite{Zheng2018b}, contains a term that requires $\mathcal{O}({d^3})$ computations. This has motivated the development of higher-efficiency acyclicity constraints for continuous optimization approaches to structure discovery, such as the one in LEAST \cite{Zhu2020LEAST}. One further way to alleviate the issues when confronted with high-dimensional problems is to encode the data into a lower-dimensional representation. This was undertaken in CausalVAE \cite{Yang2020}, who applied the NO TEARS constraint to a graph operating in low-dimensional representation space. Whilst this approach works well for non-semantic data (such as pixel data from images), it might not be useful in situations whereby the data are both high-dimensional \textit{and} semantic (as with gene regulation data in the DREAM5 dataset \cite{Marbach2012DREAM}). In the latter case, encoding semantic data into a new subspace may or may not be meaningful, and will likely depend on the domain of application.

In terms of what we consider to present the most opportunity for future work, we note that there are relatively few continuous optimization approaches which seek to learn structured, semantic representations from non-semantic, high-dimensional data such as video or image data (exceptions include CausalVAE \cite{Yang2020} and DEAR \cite{Shen2020}, and related works on scene understanding include \cite{Bear2020, Ehrhardt2020, Engelcke2019, monet}). Interestingly, the field of reinforcement learning, which involves the interaction of learning agents with each other and their environment, has been relatively slow on the uptake of causal perspectives \cite{Haan2019, Ashton2020}. Ashton (2020) \cite{Ashton2020} even notes that one of the seminal texts on reinforcement learning \cite{SuttonBarto} makes no explicit reference to causality throughout the entire text. As such, the application of causal discovery to reinforcement learning presents significant opportunity.\footnote{Some exceptions include \cite{Ross2011,Zhang2020d, Herlau2020, Rezende2020, anonICLR, Haan2019, Nair2019,Sontakke2020, Corcoll2020, Zhang2019b}.} Finally, whilst there were numerous combinatoric methods which are designed to handle unobserved confounding and/or cyclicity (\textit{e.g.}, CCD \cite{Richardson1996}, backshift \cite{Rothenhausler2015}, CCI \cite{Strobl2018}), there are relatively few such continuous optimization approaches. Given the complexity of time-varying real-world phenomena and the potential for cycles, we note the opportunity to develop continuous optimization methods which can operate in a broader class of scenarios.

\subsection{The Causal Leap}

It was mentioned in Section 1, that a causal perspective is crucial to the empirical sciences as well as for improving machine learning methods. More fundamentally, as humans we are interested in how to reason about and interact in a world full of causal interactions. In general, the pursuit of causality is essential to understanding the world and our universe. However, it is fraught with difficulty, and below we finish with a discussion on some of the criticisms and warnings relating to this otherwise laudable pursuit. 

We now take the time to discuss how structure discovery methods take us from a structural association (albeit, an association which may exhibit directional asymmetry) to that of a causal association. What is there to suggest that learning or identifying such a graphical or structural model is equivalent to learning or identifying causes and generative structure in reality? In order to interpret graphical models causally, the the Causal Markov Condition (CMC) \cite{Spirtes2000} is often assumed. However, in our view (and see also \cite{Freedman1998}) the CMC simply represents an uninformative re-branding of the regular Markov condition (which describes the conditional independence properties of the graph), with the additional and rather audacious interpretation of the arrows as directed causal dependencies. As Dawid (2008) \cite[p.83]{Dawid2008} argues, "there is no reason to believe [the causal implications of the CMC] hold in complete generality". It should be clear that the conditional independence properties of DAGs play a foundational r\^{o}le in causal discovery. However, as Dawid (2008) states in his work \textit{Beware of the DAG!}: "...for conditional independence the arrows are nothing but incidental construction features supporting the $d$-separation semantics." It may be interesting, then, to observe just how many structure discovery methods, particularly those which rely exclusively on conditional independencies (rather than, say, interventional data and rigorous identifiability), uncritically label themselves as causal... 

The use of structural equations gets us somewhat closer to where we want to be when seeking to represent causality, than do graphical models alone. This is because the structural equation formalism  can be more specific and informative than its simpler (yet intuitive) graphical counterpart \cite[p.106]{Peters2017}. Nonetheless, as with graphical models, the interpretation of structural equations as structural causal models cannot be made without strong and often untestable assumptions. Applying these strong assumptions to structural or graphical models incites some harsh criticism. Indeed, Korb \& Wallace (1997) caricature research into causal discovery as "a glorious perversion" akin to the "search for the philosopher's stone" \cite[p.551]{Korb1997}. 

Such criticisms are important to assimilate, and they remind us to be careful when using statistical/causal models to draw inference about the nature of reality. In particular, even if a graphical model bears resemblance to our own conception of a phenomenon, it may not be an appropriate or fair way to represent complex social constructs (\textit{e.g.}, gender or race), representing what Freidman described as a biased attempt to "quantify the qualitative, make discrete the continuous, or formalize the nonformal" \cite{Friedman1996}. For instance, it is not clear what it means to be able to manipulate/intervene on someone's race, independently of their other attributes, or indeed at all. In general, we need a thorough understanding of what a variable is \textit{supposed} to represent, and whether it actually represents it at all (both a problem of ontology and epistemology) before we perform meaningful inference. However, a sufficiently clear understanding may be slippery and, in some cases, impossible to attain. 

The prevalence of reports of systemic bias arising from automated decision processes is increasing, and an awareness for sources of bias is critical in undertaking fair and equitable machine learning \cite{holstein2019, nabi2019,Vowels2020, slack2019}. Just because causal discovery methods define themselves as `causal', does not mean there are not significant problems with taking the leap from data to reality. Indeed, blindly interpreting structured models as robustly representing causal quantities can be immensely problematic. We appreciate Dawid's \cite{Dawid2008} reference to Bourdieu who warns of "sliding from the model of reality to the reality of the model" \cite{Bourdieu1977}.

In spite of the notable criticisms, causal discovery methods may still be used productively, particularly for exploratory purposes (\textit{e.g.}, in providing candidate causal links for further investigation and validation) \cite{Dawid2008}. Furthermore, the combination of observational and interventional/experimental data may provide us with opportunities to uniquely \textit{identify} models which, at least under various assumptions, correspond with some true external cause-effect relationships. More broadly, shifting from naive associational and purely predictive machine learning models to models informed by causal structure, may bring concomitant improvements in model robustness and generalizability. So long as researchers maintain a cautious approach when making the leap from modelling structure to inferring causality, structure discovery methods can still be used in support of the endeavour to further human understanding.


\bibliographystyle{ACM-Reference-Format}
\bibliography{NN.bib}

\appendix

\end{document}